\documentclass[12pt]{article}

\usepackage{graphicx}
\usepackage{natbib} 
\usepackage{url} 

\newcommand{\blind}{0}

\usepackage[utf8]{inputenc}

\usepackage{caption}
\usepackage{subcaption}

\usepackage{amsmath}
\usepackage{amsfonts}
\usepackage{amssymb}
\usepackage{amsthm}
\usepackage{graphicx}
\usepackage{adjustbox}
\usepackage{hhline}
\usepackage{booktabs}
\usepackage{blindtext}
\usepackage{enumitem}
\usepackage{soul}
\usepackage{ulem}
\usepackage{bm}

\usepackage{algorithm}
\usepackage{algpseudocode}
\usepackage{physics}

\usepackage{mathrsfs}

\usepackage{mathtools}

\makeatletter
\def\BState{\State\hskip-\ALG@thistlm}
\makeatother

\usepackage{tikz}
\usepackage{pgfplots}
\pgfplotsset{compat=newest}
\usetikzlibrary{shapes.geometric,arrows,fit,matrix,positioning,patterns}
\tikzset
{
    treenode/.style = {ellipse, draw=black, align=center, minimum size=1cm},
    subtree/.style  = {isosceles triangle, draw=black, align=center, minimum height=0.5cm, minimum width=1cm, shape border rotate=90, anchor=north},
    terminal/.style  = {rectangle, draw=black, align=center, minimum height=0.8cm, minimum width=1cm, anchor=north}
}

\newtheorem{thm}{Theorem}

\newtheorem{defin}{Definition}
\newtheorem{example}{Example}

\usepackage{todonotes}   
\usepackage{lipsum}

\newcommand{\x}{\mathbf{x}}
\newcommand{\X}{\mathbf{X}}
\newcommand{\y}{\mathbf{y}}

\newcommand{\Y}{\mathbf{Y}}

\newcommand{\Var}{\texttt{Var}}

\newcommand{\1}{\mathbb{I}}
\newcommand{\R}{\mathbf{R}}
\newcommand{\E}{\mathbb{E}}

\newcommand{\ens}{\mathcal{E}}
\newcommand{\Ens}{\bm{\ens}}
\newcommand{\T}{\mathcal{T}}
\newcommand{\M}{\mathcal{M}}

\newcommand{\ptrue}{\mathcal{P}}
\newcommand{\phat}{\widehat{\mathcal{P}}}

\newcommand{\pftrue}{\mathcal{P}_{\mathcal{F}}}
\newcommand{\pfhat}{\widehat{\mathcal{P}_{\mathcal{F}}}}
\newcommand{\pstrue}{\mathcal{P}_{\mathcal{X}}}
\newcommand{\pshat}{\widehat{\mathcal{P}_{\mathcal{X}}}}

\newcommand{\iset}{\mathcal{X}}
\newcommand{\oset}{\mathcal{Y}}

\newcommand{\f}{\mathbf{f}}

\newcommand{\A}{\mathcal{A}}

\newcommand{\iid}{\stackrel{iid}{\sim}}
\makeatletter
\DeclareFontEncoding{LS1}{}{}
\DeclareFontSubstitution{LS1}{stix}{m}{n}
\DeclareMathAlphabet{\mathscr}{LS1}{stixscr}{m}{n}
\makeatother

\newcommand{\dat}{\mathscr{D}}

\usepackage[nodisplayskipstretch]{setspace}
\usepackage{csquotes}
\usepackage{tikz-cd} 
\usepackage{hyperref}
\usepackage{array,multirow}
\usepackage[affil-it]{authblk}

\begin{document}

\def\spacingset#1{\renewcommand{\baselinestretch}%
{#1}\small\normalsize} \spacingset{1}


\if0\blind
{
  \title{\bf Using BART to Perform Pareto Optimization and Quantify its Uncertainties}
    \author[1]{Akira Horiguchi\thanks{Corresponding author. Email: \texttt{horiguchi.6@osu.edu}}}
    \author[1]{Thomas J. Santner}
    \author[2]{Ying Sun}
    \author[1]{Matthew T. Pratola}
    \affil[1]{Department of Statistics \protect\\ 
    The Ohio State University \protect\\ 
    Cockins Hall \protect\\ 
    1958 Neil Ave. \protect\\ 
    Columbus, OH 43210}
    \affil[2]{Statistics Program \protect\\ 
    King Abdullah University of Science and Technology \protect\\ 
    Thuwal 23955-6900 \protect\\ 
    Saudi Arabia}
  \maketitle
} \fi


\if1\blind
{
  \bigskip
  \bigskip
  \bigskip
  \begin{center}
    {\LARGE\bf Using BART to Perform Pareto Optimization and Quantify its Uncertainties}
\end{center}
  \medskip
} \fi

\bigskip
\begin{abstract}
Techniques to reduce the energy burden of an industrial ecosystem often require solving a multiobjective optimization problem. 
However, collecting experimental data can often be either expensive or time-consuming.
In such cases, statistical methods can be helpful.
This article proposes Pareto Front (PF) and Pareto Set (PS) estimation methods using Bayesian Additive Regression Trees (BART), which is a non-parametric model whose assumptions are typically less restrictive than popular alternatives, such as Gaussian Processes (GPs). 
These less restrictive assumptions allow BART to handle scenarios (e.g. high-dimensional input spaces, nonsmooth responses, large datasets) that GPs find difficult.
The performance of our BART-based method is compared to a GP-based method using analytic test functions, demonstrating convincing advantages.
Finally, our BART-based methodology is applied to a motivating engineering problem.
Supplementary materials, which include a theorem proof, algorithms, and R code, for this article are available online.
\end{abstract}

\noindent%
\vfill


\newpage

\section{Introduction}
\label{sec:intro}

Many important Industry 4.0 problems \citep[e.g.][]{Xie2014,Han2016,Ivanov2016} can be formulated as multiobjective optimization (MO) problems.  
For example, \cite{Shukla2020} describe how the use of dynamic voltage scaling in real-time embedded systems (RTES) produces the two mutually conflicting objectives of energy efficiency and timeliness of task execution.
Furthermore, the timing constraints of tasks in RTES can only be approximated, which prompts the need to quantify this imprecision. 

More generally, if each objective in a problem corresponds to an output dimension in the vector-valued function $\f(\bm{\cdot})$, the goal of MO is to ``minimize'' (or ``maximize'', depending on the application) this function. 
Seldom will all of these objectives be simultaneously minimized by the same input setting.
Hence, we seek to find the set of best compromises between competing objectives and the set of all inputs that produce these compromises.
(Section \ref{sec:mo} provides a mathematical definition of best compromises.)
The former set is called the Pareto Front (PF); the latter set is called the Pareto Set (PS). 

When the function cannot be explicitly evaluated or where the number of evaluations is limited, statistical methods can be helpful.
A common strategy in computer experiments for PF and PS estimation is to approximate $\f(\bm{\cdot})$ by a surrogate model trained on a small number of evaluated points and perform inference on this fitted surrogate model \citep[e.g. see][]{Svenson11}.  
\cite{Binois15} achieve PF estimation and uncertainty quantification (UQ) by fitting a Gaussian Process (GP), simulating approximate realizations of the fitted GP, and treating the resulting approximate conditional PFs as random sets. 
However, their second step requires discretizing the input space into a finite number of points, which may be computationally expensive if the input dimension is large. 
Furthermore, no work in general has been done to quantify the uncertainty of PS estimation with GPs. 

A popular alternative to the GP for emulating single-output simulators is the Bayesian Additive Regression Trees (BART) model introduced by \cite{CGM10} (CGM). 
BART partitions the input space into hyperrectangles and applies a constant mean model to each hyperrectangle.
Unlike GP, BART can capture nonstationarity, avoids $O(n^3)$ matrix decompositions during fitting, easily handles categorical inputs, and typically has fewer restrictive assumptions, which makes BART feasible in a wider range of scenarios if enough training samples are provided. 
In particular, BART is well-suited to problems with large input dimensions and large datasets \citep{Pratola14}. 
\cite{Breiman01}'s Random-Forest model is also used in surrogate-based optimization and retains many of BART's advantages over GP, but lacks BART's natural UQ capabilities via its Bayesian formulation.

Though BART has been used for single-objective optimization \citep[e.g. in][]{Chipman12}, it has never been used to perform multiobjective optimization. 
Our primary contribution is twofold. 
We first find the PF and PS of exact simulated realizations of a fitted multiple-output BART model and hence avoid grid approximations of the input space. 
We then quantify the uncertainty of these estimates of the PF and PS of $\f(\bm{\cdot})$ using random sets \citep{Binois15} and our novel extension of the depth approach described in \cite{LopezPintado09,SGN12,WMK13}. 

The paper is organized as follows. 
Section \ref{sec:bart} introduces BART with multiple outputs. 
Given a multiple-output BART function, Section \ref{sec:mo} establishes how to find its image, PF, and PS.
In Section \ref{sec:uq}, we derive UQ measures for BART-based PF and PS estimates. 
In Section \ref{sec:simstudy}, we perform simulation studies, comparing our approach to the popular GP approach. 
Section \ref{sec:engapp} demonstrates our BART-based methodology on an engineering application.
Section \ref{sec:conc} concludes the paper with a discussion. 
Proofs of stated theorems can be found in the Supplementary Materials.

\section{BART}
\label{sec:bart}

We observe data $\dat \coloneqq \{(\y(\x_i), \x_i)\}_{i=1}^n$.
Each output $\y(\x_i) = (y_1(\x_i), y_2(\x_i), \ldots, y_d(\x_i)) \in \mathbb{R}^d$ is assumed to be a realization of the random variable 
\begin{equation}
\label{eq:datagen} 
\Y(\x_i) = \f(\x_i) + \bm{\epsilon}_i, \quad
\f(\bm{\cdot}) = (f_1(\bm{\cdot}), f_2(\bm{\cdot}), \ldots, f_d(\bm{\cdot})) \colon \iset \rightarrow \mathbb{R}^d
\end{equation}
where $\f(\bm{\cdot})$ is the vector-valued function described in Section \ref{sec:intro}, 
each $y_j(\bm{\cdot})$ and $f_j(\bm{\cdot})$ has common domain $\iset \subset \mathbb{R}^p$, noise vectors $\bm{\epsilon}_1, \ldots, \bm{\epsilon}_n \iid N_d(\bm{0}, \bm{\sigma}^2 I_d)$, and parameter $\bm{\sigma}^2 = (\sigma^2_1, \ldots, \sigma^2_d)$. 
We assume the domain $\iset$ is a $p-$dimensional bounded hyperrectangle.

\subsection{Multiple-output BART}
\label{sec:bartmultipleoutput}

To make inference on the unknown $\f(\bm{\cdot})$, we approximate each marginal $f_j(\bm{\cdot})$ (for $j = 1, \ldots, d$) by fitting a BART model to the marginal data set $\dat_j \coloneqq \{(y_j(\x_i), \x_i)\}_{i=1}^n$. 
These $d$ independently fitted BART models define our $d-$output BART model: 
the $i$th posterior draw of the $d-$output BART model is $\bm{\Theta}^{(i)} = \left(\Theta_1^{(i)}, \ldots, \Theta_d^{(i)}\right)$, where each $\Theta_j^{(i)}$  is the $i$th posterior draw of the $j$th fitted BART model.

Section 2 of \cite{Horiguchi21} describes how a posterior draw $\Theta$ of a BART model induces a regression function $\ens(\bm{\cdot}; \theta)$, where $\theta \coloneqq \{(\T_t, \M_t)\}_{t=1}^m$.
Similarly, each posterior draw $\bm{\Theta}$ of our $d-$output BART model induces the $d-$output regression function 
\begin{equation}
\Ens(\bm{\cdot}; \bm{\theta}) = (\ens(\bm{\cdot}; \theta_1), \ens(\bm{\cdot}; \theta_2), \ldots, \ens(\bm{\cdot}; \theta_d)) \colon \iset \rightarrow \mathbb{R}^d
\label{eq:multibart}
\end{equation}
where $\bm{\theta} = (\theta_1, \ldots, \theta_d)$ and each $\theta_j$ comes from the $j$th fitted BART model.

\subsection{Prior specification}
\label{sec:bartprior}


Here we describe our prior specifications for tree-topology parameters $\T_t$, leaf-node parameters $\mu_{tk}|\T_t$, and noise variance $\sigma^2$ because we ultimately deviate from CGM's default hyperparameter values for our multiobjective optimization problem. 

The $\pi(\T_t)$ prior decomposes into three components: tree depth, split variable at each internal node, and cutpoint value at each internal node. 
We leave details of the first two components to CGM and \cite{CGM98}. 
For the cutpoint value of any given split variable, this paper uses a discrete uniform prior of $30$ values over the range of the observed input values. 

CGM model $\mu_{tk}|\T_t$ with a Gaussian prior $N(0, (4\kappa^2m)^{-1})$ (after centering and rescaling the output data so that the minimum and maximum observed transformed response values are, respectively, $y_{min} = -0.5$ and $y_{max} = 0.5$). 
Under the sum-of-trees model, the prior on $\E[Y(\x)]$ then becomes $N\left(0, (4\kappa^2)^{-1}\right)$, where CGM default to $\kappa=2$.
For single-objective optimization, however, \cite{Chipman12} use $\kappa=1$ to allow BART to produce more pronounced optima. 
For similar reasons, we use $\kappa=1$ and $m=30$ trees for our applications in Section \ref{sec:simstudy}, but other situations may call for different $(\kappa, m)$ values.
We also set the minimum number of observations allowed in each leaf node to ten. 

For $\pi(\sigma^2)$, we use the scaled inverse chi-square distribution $\sigma^2 \sim \text{Scale-}\chi^{-2}(\nu, \lambda)$ with values $\nu=3$ and $\lambda = 0.01^2$ chosen to induce a prior mean of $0.0003$ (see CGM for details of $\nu$ and $\lambda$ selection). 
However, we find that the hyperparameter $\kappa$ more strongly influences the smoothness of the response.

\section{Multiobjective optimization}
\label{sec:mo}

This section details how we can find the PF and PS of the multiple-output BART regression function \eqref{eq:multibart} given some fixed $\bm{\theta}$.
However, we must first introduce the notion of Pareto dominance,
which we use to identify the best compromises between competing objectives.

\begin{figure}[t!]
    \centering
    \begin{subfigure}[t]{0.31\textwidth}
        \centering
        \includegraphics[width=\textwidth]{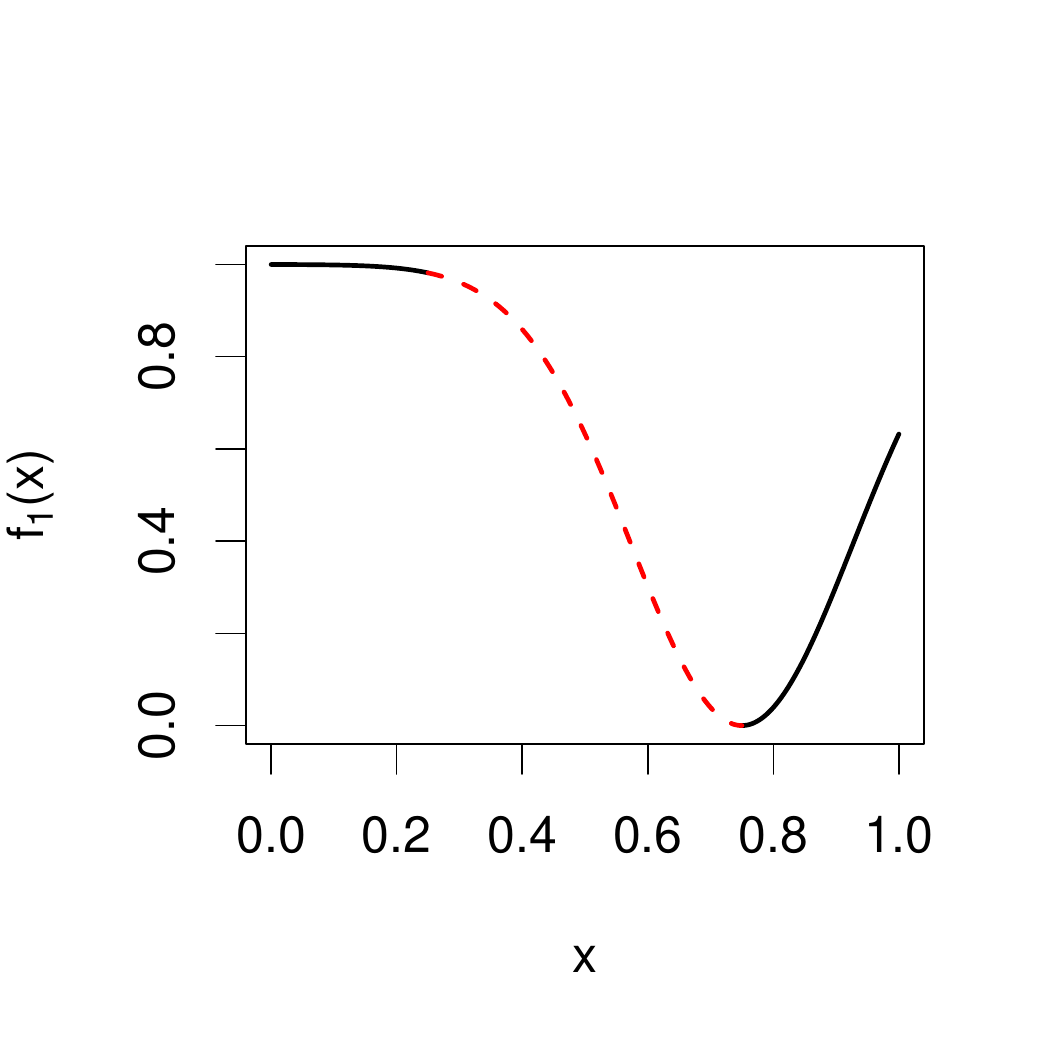}
        \caption{$f_1(x)$ vs $x$.}
        \label{fig:f1}
    \end{subfigure}
    ~
    \begin{subfigure}[t]{0.31\textwidth}
        \centering
        \includegraphics[width=\textwidth]{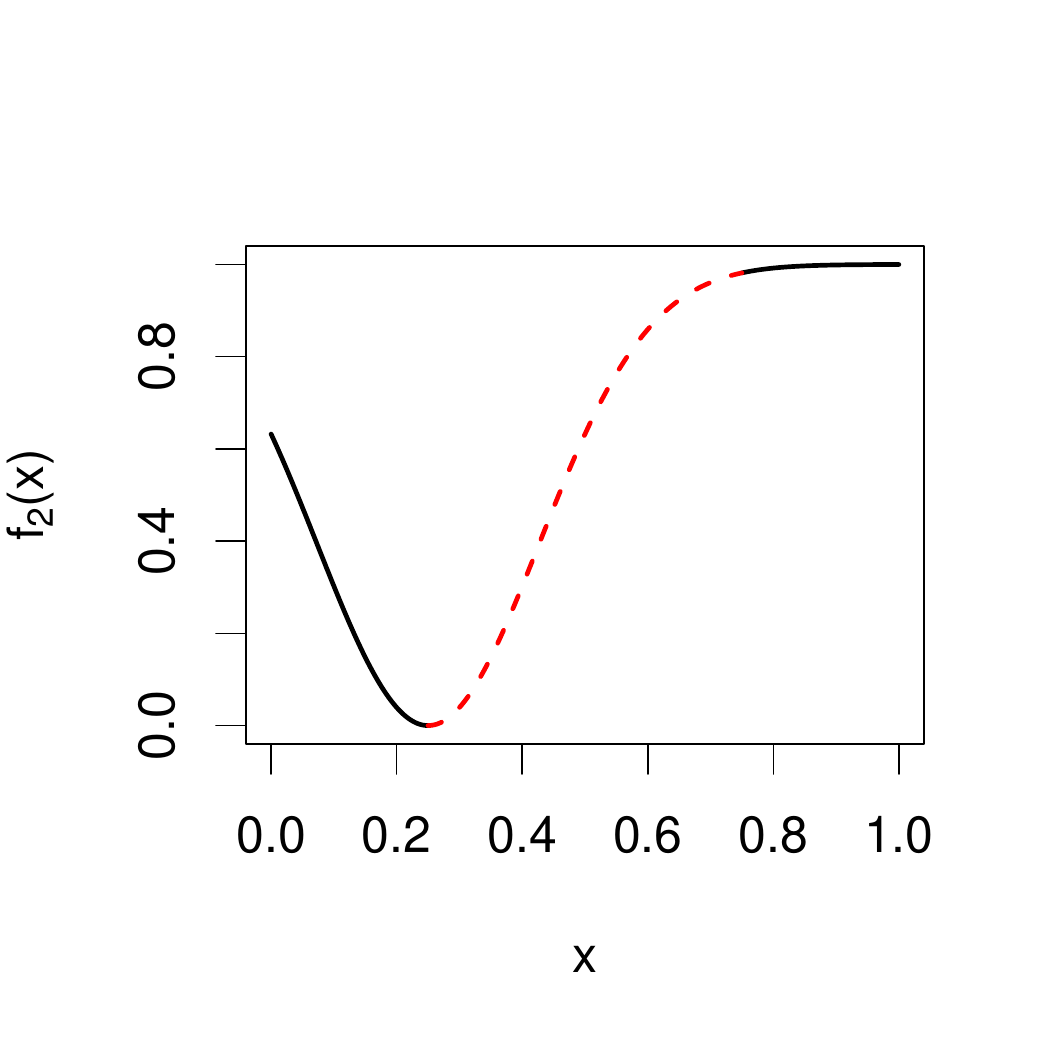}
        \caption{$f_2(x)$ vs $x$.}
        \label{fig:f2}
    \end{subfigure}
    ~
    \begin{subfigure}[t]{0.31\textwidth}
        \centering
        \includegraphics[width=\textwidth]{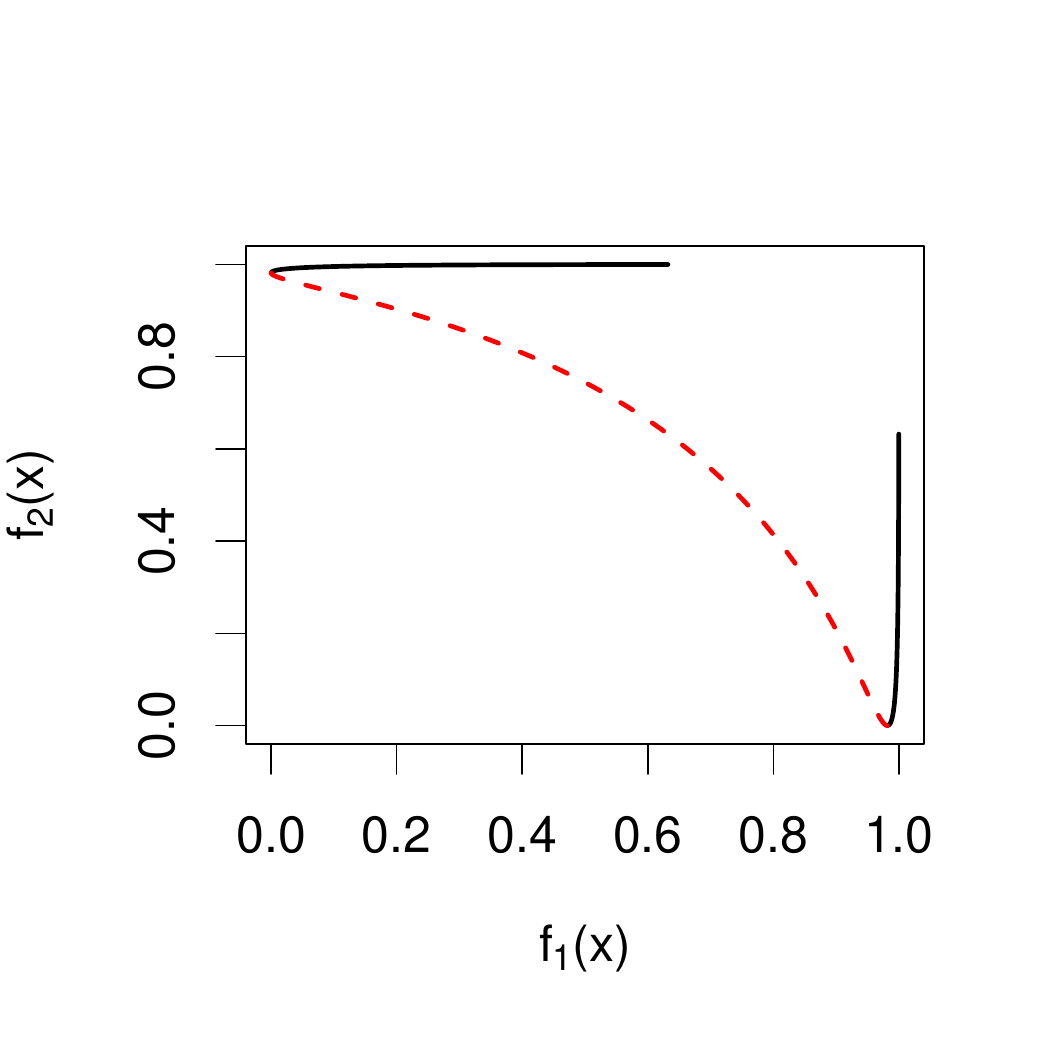}
        \caption{$f_2(x)$ vs $f_1(x)$.}
        \label{fig:f1f2}
    \end{subfigure}
    \caption{A biobjective function $\f(x) = (f_1(x), f_2(x))$ where $x \in [0, 1]$.}
    \label{fig:appmo}
\end{figure}

\begin{defin}[Pareto dominance]
The objective point $\mathbf{v} = (v_1, \ldots, v_d) \in \mathbb{R}^d$ (weakly) dominates the point $\mathbf{w} = (w_1, \ldots, w_d) \in \mathbb{R}^d$ (denoted $\mathbf{v} \succeq \mathbf{w}$) if $v_j \leq w_j$ for all $j = 1, \ldots, d$.
If at least one of these inequalities is strict, we say $\mathbf{v}$ strictly dominates $\mathbf{w}$ (denoted $\mathbf{v} \succ \mathbf{w}$).
\end{defin}
We can now precisely define a multiobjective function's PF and PS:
the PF is the set of all nondominated image points; the PS is the set of all inputs that produce the PF.
For example, consider Figure \ref{fig:f1f2}, which shows the image of a biobjective function.
Any point on the dashed segment is not dominated by any other image point while any image point on a solid segment is dominated by at least one other image point.
Thus, the PF is the dashed segment in Figure \ref{fig:f1f2} and the PS is the interval $[0.25, 0.75]$ in Figures \ref{fig:f1} and \ref{fig:f2}.

We find the PF and PS of \eqref{eq:multibart} using the two-step algorithm:
\begin{enumerate}
    \item (Section \ref{sec:bartimg}) Find the image $\Ens(\iset; \bm{\theta})$ and corresponding input hyperrectangles.
    \item (Section \ref{sec:bartpareto}) Find all nondominated points in the image $\Ens(\iset; \bm{\theta})$. 
\end{enumerate}
Step 2 produces the desired PF.
The desired PS is the union of all input hyperrectangles (found in Step 1) that yields a point in the PF. 

\subsection{Finding the image of a multiple-output BART function}
\label{sec:bartimg}

The following example will make it clear that \eqref{eq:multibart} has a finite image.
We can find these image points and their corresponding input hyperrectangles using the parameter values $\bm{\theta}$.

\begin{figure}[ht]
    \centering
    \begin{subfigure}[b]{\textwidth}
        \centering
        \begin{equation*}
        \left(
        \begin{array}{c}\displaystyle%
        \begin{tikzpicture}[->,>=stealth', level/.style={sibling distance = 3cm/#1, level distance = 1.3cm}, scale=0.8,transform shape]
            \node [treenode] {$x_1 < 0.3$}
            child { node [treenode] {$x_2 < 0.4$}
                  child { node [terminal, dashed, red] {$-3$} }
                  child { node [terminal] {$3$} } 
                  }
            child { node [terminal] {$1$} }
        ;
        \end{tikzpicture}
        \quad
        \raisebox{35pt}{\pmb{+}}
        \quad
        \begin{tikzpicture}[->,>=stealth', level/.style={sibling distance = 1.5cm/#1, level distance = 1.3cm}, scale=0.8,transform shape]
            \node [treenode] {$x_1 < 0.6$}
            child { node [terminal, dashed, red] {$-2$} }
            child { node [terminal] {$2$} }
        ;
        \end{tikzpicture}
        \quad \quad
        {\Large \pmb{,}}
        \quad \quad
        \begin{tikzpicture}[->,>=stealth', level/.style={sibling distance = 3cm/#1, level distance = 1.3cm}, scale=0.8,transform shape]
            \node [treenode] {$x_2 < 0.2$}
            child { node [treenode] {$x_1 < 0.1$}
                  child { node [terminal, dashed, red] {$-1$} }
                  child { node [terminal] {$1$} } 
                  }
            child { node [terminal] {$2$} }
        ;
        \end{tikzpicture}
        \quad
        \raisebox{35pt}{\pmb{+}}
        \quad
        \begin{tikzpicture}[->,>=stealth', level/.style={sibling distance = 1.5cm/#1, level distance = 1.3cm}, scale=0.8,transform shape]
            \node [treenode] {$x_1 < 0.5$}
            child { node [terminal, dashed, red] {$-3$} }
            child { node [terminal] {$3$} }
        ;
        \end{tikzpicture}
        \end{array}
        \right)
        \end{equation*}
        \vspace{5mm}
    \end{subfigure}
    
    \begin{subfigure}[b]{0.4\textwidth}
      \begin{tikzpicture}[scale=0.85]
          \begin{axis}[xmin=0, xmax=1, ymin=0, ymax=1, xlabel={$x_1$}, ylabel={$x_2$}]
              \addplot [dotted, thick] table {
                  0.3 0.0
                  0.3 1.0
              };
              \addplot [dotted, thick] table {
                  0.6 0.0
                  0.6 1.0
              };
              \addplot [dotted, thick] table {
                  0.5 0.0
                  0.5 1.0
              };
              \addplot [dotted, thick] table {
                  0.0 0.2
                  1.0 0.2
              };
              \addplot [dotted, thick] table {
                  0.0 0.4
                  0.3 0.4
              };
              \addplot [dotted, thick] table {
                  0.1 0.0
                  0.1 0.2
              };
              \filldraw[pattern=north west lines, pattern color=red, draw=none] (0,0) rectangle (.1,.2);
          \end{axis}
      \end{tikzpicture}
    \end{subfigure}
    \hspace{15mm}
    \begin{subfigure}[b]{0.4\textwidth}
        \begin{tikzpicture}[scale=0.85]
            \begin{axis}[xmin=-7, xmax=5, ymin=-5.5, ymax=7, xlabel={$\ens(\bm{\cdot}; \theta_1)$}, ylabel={$\ens(\bm{\cdot}; \theta_2)$}]
                \addplot [only marks, ultra thick, mark size=3pt] table {
                -5 -2
                -1 -2
                -1  4
                 3  4
                 1 -1
                -1 -1
                -1  5
                 3  5
                -5 -1
                };
                \addplot [only marks, ultra thick, dashed, red, mark=triangle, mark size=7pt] table {
                -5 -4
                };
            \end{axis}
        \end{tikzpicture}
    \end{subfigure}
    \caption{A biobjective function $\Ens(\bm{\cdot}; \bm{\theta}) = (\ens(\bm{\cdot}; \theta_1), \ens(\bm{\cdot}; \theta_2))$ with tree ensembles (top row) with parameters $\theta_1$ (left) and $\theta_2$ (right). Bottom left: input space partitioned into 10 rectangles. Bottom right: output space with all 10 image points of $\Ens(\bm{\cdot}; \bm{\theta})$.}
    \label{fig:exmooens1}
\end{figure}
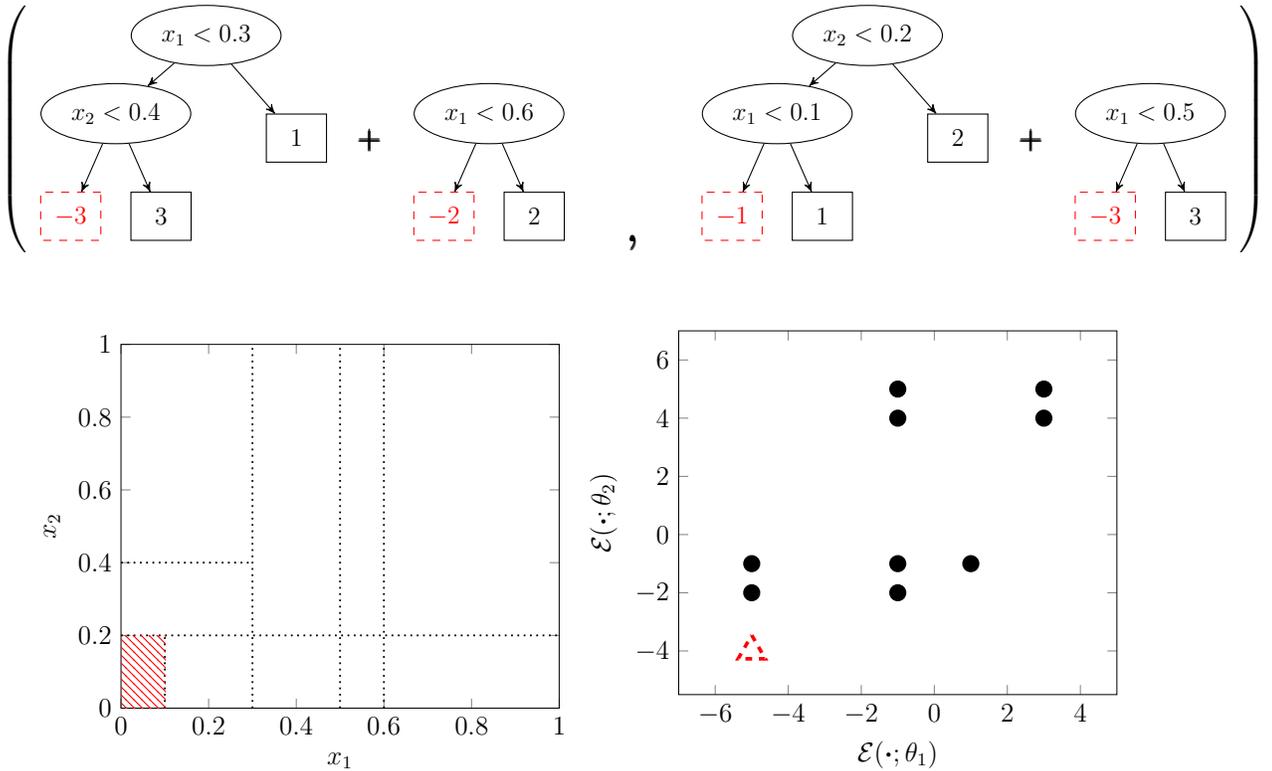

\begin{example}
\label{ex:bifun}
\upshape
Here, we find the image of the biobjective function $\Ens(\bm{\cdot}; \bm{\theta})$ shown in Figure \ref{fig:exmooens1}, where $\iset = [0, 1]^2$.  
Any input belongs to one leaf node per tree; these $dm=4$ leaf nodes correspond to an image point.
For example, the input $\x^* = (0.1, 0.07)$ belongs to the four dash-bordered leaf nodes, which correspond to the image point $\Ens(\x^*; \bm{\theta}) = (-5, -4)$ shown as a dashed triangle.
Conversely, this image point corresponds to the hatch-filled input rectangle, which is the set of all inputs that belong to all four dash-bordered leaf nodes.

Of the $(3\times2) (3\times2) = 36$ possible one-leaf-node-per-tree combinations, only $10$ are valid and produce an image point. 
For an example of an invalid combination, an input point cannot belong to both the right and left leaf nodes of, respectively, the first and second ensemble's right tree; 
the conditions $x_1 \geq 0.6$ and $x_1 < 0.5$ cannot be simultaneously satisfied.
With only $10$ valid combinations, the function $\Ens(\bm{\cdot}; \bm{\theta})$ has only $10$ image points (shown in the bottom right plot of Figure \ref{fig:exmooens1}). 

Between the two ensembles, the six split rules together partition the input space such that the resulting set of partitioning rectangles (shown in the bottom left plot of Figure \ref{fig:exmooens1}) is bijective to the set of valid one-leaf-node-per-tree combinations, which is itself bijective to the image of $\Ens(\bm{\cdot}; \bm{\theta})$.
Thus, the function $\Ens(\bm{\cdot}; \bm{\theta})$ can be written as a linear combination of indicator functions each corresponding to a partitioning rectangle.
\hfill $\blacksquare$
\end{example}

Theorem \ref{thm:multibartimg} describes a similar result but for any $d$-output BART function $\Ens(\bm{\cdot}; \bm{\theta})$:
if we obtain every possible one-leaf-node-per-tree combination, we can find the image and corresponding input (hyper)rectangles of the $d$-objective function. 
The proof of Theorem \ref{thm:multibartimg} (see Supplement) provides more insight into how these two tasks can be achieved.

\begin{thm}
\label{thm:multibartimg}
Any $d$-output BART function in the form of \eqref{eq:multibart} can be written as a linear combination of indicator functions of hyperrectangles: 
\begin{equation*}
\Ens(\bm{\cdot}; \bm{\theta}) = \sum_{q \in B_{\Ens}} \bm{\alpha}_q \1_{\R_q}(\bm{\cdot}),
\end{equation*}
where the set $B_{\Ens}$ indexes the valid one-leaf-node-per-tree combinations in $\bm{\theta}$, each $\bm{\alpha}_q \in \mathbb{R}^d$ is an image point of $\Ens(\bm{\cdot}; \bm{\theta})$, and the set of hyperrectangles $\{\R_q\}_{q \in B_{\Ens}}$ partitions $\iset$.
\end{thm}

\subsection{Finding the PF and PS of multiple-output BART function}
\label{sec:bartpareto}

After finding the image of a $d-$output BART function $\Ens(\bm{\cdot}; \bm{\theta})$, we find its set of nondominated points using an efficient recursive algorithm from \cite{Kung75}.
For simplicity, we describe (in the Supplement) only one of these algorithms, which finds the nondominated points in a finite set $V$ of $d-$dimensional vectors.
In our setting, the set $V$ is the image of $\Ens(\bm{\cdot}; \bm{\theta})$. 
\cite{Kung75} shows that the algorithm's time complexity has an upper bound of $O(|V| (\log_2 |V|)^{d-2})$ if $d>3$.
This algorithm is still valid if $d \in \{2,3\}$, but this upper bound no longer applies in these cases. 

The desired PS is the union of all input hyperrectangles corresponding to the nondominated image points.
For the example in Figure \ref{fig:exmooens1}, the PF is the dashed triangle image point while the PS is the hatched input rectangle.

\section{Uncertainty quantification}
\label{sec:uq}

\begin{figure}[h]
\centering
\begin{tikzpicture}[scale=0.9]
    \begin{axis}[xmin=0, xmax=1, ymin=0, ymax=1, xlabel={$y_1$}, ylabel={$y_2$}]
        \addplot [only marks, blue, mark=square*, mark size=3pt] table {
        0.0 1.0
        0.05 0.4
        0.6 0.0
        };
        \filldraw[pattern=north west lines, pattern color=blue, draw=none] (0.05,0.4) rectangle (1,1);
        \filldraw[pattern=north west lines, pattern color=blue, draw=none] (0.6,0.0) rectangle (1,0.4);
        \addplot [only marks, orange, mark=triangle*, mark size=3.5pt] table {
        0.4 0.2
        1.0 0.0
        };
        \filldraw[pattern=north east lines, pattern color=orange, draw=none] (0.4,0.2) rectangle (1,1);
    \end{axis}
\end{tikzpicture}
\caption{Example of 2 CPFs (triangles and squares) and the corresponding DPSCs (areas with hatched lines) in objective space $\oset=[0,1]^2$.}
\label{fig:dpsc}
\end{figure}
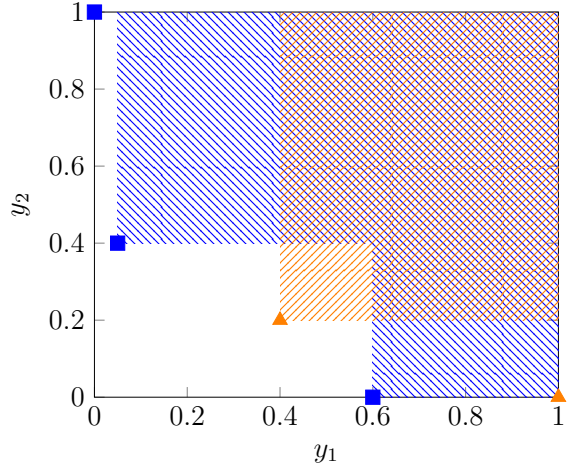

We can quantify the uncertainty of PF estimates induced by a fitted multiple-output BART model by
making $N$ independent draws from the posterior,
creating for each draw the resulting BART regression function as defined in \eqref{eq:multibart}, 
and finding each conditional PF (CPF) as described in Section \ref{sec:mo}.
This section details two UQ approaches using the sample of $N$ CPFs, which we denote as $c^{(1)}, \ldots, c^{(N)}$. 
Both approaches use dominated point set closures (DPSCs): for $i=1, \ldots, N$, define the DPSC $A^{(i)}$ of a CPF $c^{(i)}$ to be the closure of the set of points dominated by at least one point in $c^{(i)}$.
That is, $A^{(i)} \coloneqq \{\y' \in \oset \mid \y' \preceq \y \text{ for at least one } \y \in c^{(i)}\}$, where $\oset \subset \mathbb{R}^d$ is the smallest compact hyperrectangle that contains every objective point in the training set $\dat$. 
Figure \ref{fig:dpsc} shows examples of DPSCs.

\subsection{Random-sets approach}

\cite{daFonseca2010} treat each DPSC as a realization of an \emph{attained set} -- a random closed set whose probability distribution is characterized by its \emph{attainment function}.
\begin{defin}[Attained set and attainment function]
The DPSC of a random PF is the set attained by the random PF.
The attainment function $\alpha_{\A} \colon \oset \rightarrow [0, 1]$ of an attained set $\A$ is defined as $\alpha_{\A}(\y) \coloneqq \mathbb{P}(\y \in \A)$ for every point $\y \in \oset$. 
\end{defin}

\begin{example}
\upshape
Consider Figure \ref{fig:dpsc} and let $A^{\color{blue}{\blacksquare}}$ ($A^{\color{orange}{\blacktriangle}}$) be the DPSC created from the square (triangle) points.
Define an attained set $\A$ to be $A^{\color{blue}{\blacksquare}}$ with probability $0.1$ and $A^{\color{orange}{\blacktriangle}}$ otherwise.
The attainment function $\alpha_{\A}$ of $\A$ is then 
\[ 
\alpha_{\A}(\y) = 
\begin{cases} 
    0,   & \y \notin A^{\color{blue}{\blacksquare}} \cup A^{\color{orange}{\blacktriangle}} \\[-15pt]
    0.1,   & \y \in A^{\color{blue}{\blacksquare}} \setminus A^{\color{orange}{\blacktriangle}} \\[-15pt]
    0.9,   & \y \in A^{\color{orange}{\blacktriangle}} \setminus A^{\color{blue}{\blacksquare}} \\[-15pt]
    1,   & \y \in A^{\color{blue}{\blacksquare}} \cap A^{\color{orange}{\blacktriangle}} 
\end{cases}
\]
which we can interpret as the probability of $\y$ being in the random set $\A$.
\hfill $\blacksquare$
\end{example}

\subsubsection{PF and PS estimation}
\label{sec:rspf}


If an attained set $\A$ has uncountably many possible set realizations, its attainment function may be difficult to formulate. 
We may instead estimate the attainment function using an empirical version, which takes on values in the set $\{0, \frac{1}{N}, \frac{2}{N}, \ldots, \frac{N-1}{N}, 1\}$. 

\begin{defin}[Empirical Attainment Function]
Let $A^{(1)}, A^{(2)}, \ldots, A^{(N)}$ be realizations of the attained set $\A$ on $\oset$. 
The empirical attainment function $\hat{\alpha}_N \colon \oset \rightarrow [0, 1]$ is defined to be the fraction of attained set realizations that contain its argument:
$\hat{\alpha}_N(\bm{\cdot}) \coloneqq \frac{1}{N} \sum_{i=1}^N \1_{A^{(i)}}(\bm{\cdot})$.
\end{defin}

With the empirical attainment function, we can quantify the uncertainty of the CPFs using an $\alpha_{RS}\%$ (where $0 < \alpha_{RS} < 1$) PF UQ point cloud, which we define to be the set of all CPF points 
$\y$ such that $\hat{\alpha}_N(\y) \in \left(0.5 - \alpha_{RS}/2, 0.5 + \alpha_{RS}/2\right)$.
That is, this UQ point cloud is the set of CPF points dominated by some proportion of CPFs, where this proportion is between $\left(0.5 - \alpha_{RS}/2, 0.5 + \alpha_{RS}/2\right)$. 
Obtaining this point cloud requires evaluating the function $\hat{\alpha}_N(\bm{\cdot})$ at every CPF point, which means checking the condition ``$\y \in A^{(i)}$'' for all CPF points $\y$ and all $i=1, \ldots, N$.
A single ``$\y \in A^{(i)}$'' check assesses $\y$'s dominance relationship with possibly every point in the CPF $c^{(i)}$ and thus
takes $\mathcal{O}(d |c^{(i)}|)$ time.
Checking ``$\y \in A^{(i)}$'' for all CPF points $\y$ and all $i=1, \ldots, N$ then takes $\mathcal{O}\big(d (\sum_{i=1}^N |c^{(i)}|)^2\big)$ time.


Regarding PS estimation, recall from Theorem \ref{thm:multibartimg} that each image point of a $d-$output BART function $\Ens(\bm{\cdot}; \bm{\theta})$ corresponds to a partitioning hyperrectangle. 
Hence, the PS of $\Ens(\bm{\cdot}; \bm{\theta})$ corresponds to a collection of hyperrectangles. 
This paper thus quantifies the uncertainty of conditional PSs to be the union of the hyperrectangles corresponding to the points in the PF UQ point cloud: 
$\bm{\pshat} \coloneqq \bigcup_{\y \in \bm{\pfhat}} \Ens^{-1}(\y; \bm{\theta})$, where $\bm{\pfhat}$ is a PF UQ point cloud and $\Ens^{-1}(\y; \bm{\theta})$ is the hyperrectangle corresponding to the objective point $\y$.

\subsection{Band depth approach}
\label{sec:banddepthapproach}

\begin{figure}[h]
    \centering
    \begin{subfigure}[t]{0.41\textwidth}
        \includegraphics[width=\textwidth]{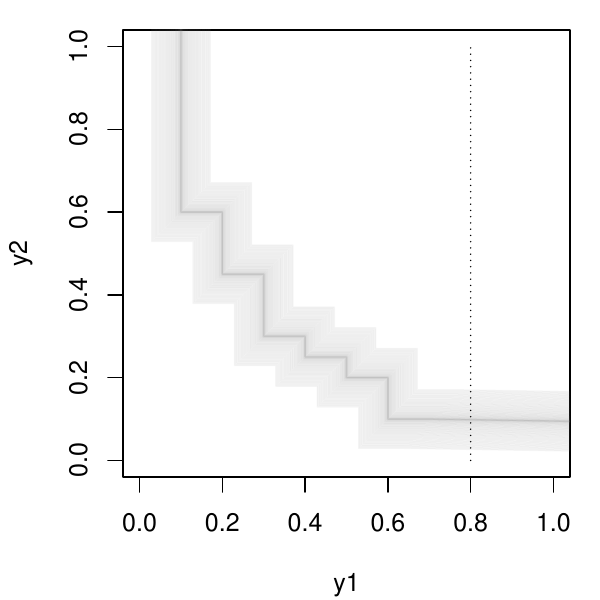}
    \end{subfigure}
    ~
    \begin{subfigure}[t]{0.41\textwidth}
        \includegraphics[width=\textwidth]{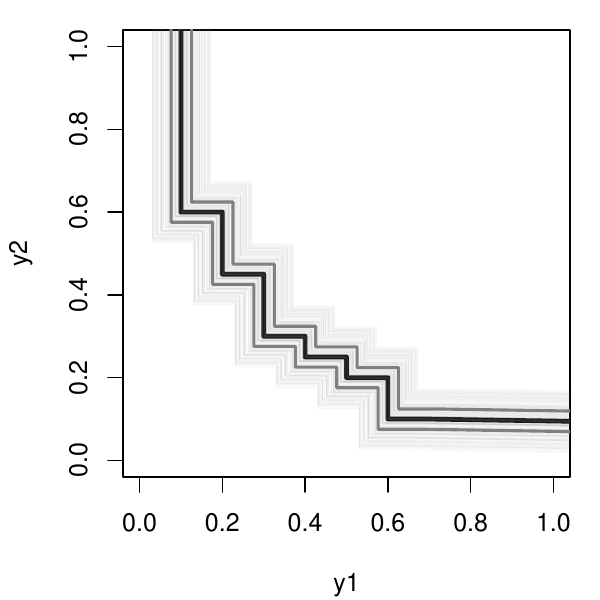}
    \end{subfigure}
    \caption{The lower-left boundaries of $N=101$ DPSCs. Left: faint gray curves correspond to the $N$ CPFs. Right: thick black curve corresponds to the deepest CPF. Thick gray curves correspond to the $50\%$ deepest CPFs corresponding to $\alpha_{MBD} = 0.5$.}
    \label{fig:exampledepth}
\end{figure}

A second approach to quantify the variability of CPFs $c^{(1)}, c^{(2)}, \ldots, c^{(N)}$ (with associated DPSCs $A^{(1)}, A^{(2)}, \ldots, A^{(N)}$) is to order them using a graph-based notion of depth. 
The idea is to measure the centrality of a curve with respect to either a set of curves or a population distribution. 
A sample of curves can then be ordered from the center outward, where the ``deepest'' curve would be the ``median'' curve. 
\cite{LopezPintado09} introduce the concept of band depth for univariate functions. 
\cite{WMK13} generalize this band depth definition to operate on sets, which we use to order $c^{(1)}, c^{(2)}, \ldots, c^{(N)}$. 
We say that a CPF $c^{(i)}$ lies in the band delimited by two CPFs $c^{(j)}$ and $c^{(k)}$ if and only if $\Big[ A^{(j)} \cap A^{(k)} \Big] \subseteq A^{(i)} \subseteq \Big[ A^{(j)} \cup A^{(k)} \Big]$.
We denote this relationship by $c^{(i)} \subseteq^* B(c^{(j)}, c^{(k)})$.
Figure \ref{fig:exampledepth} shows an example of a band delimited by two CPFs. 
We now define the band depth of $c^{(i)}$ to be the proportion of bands delimited by two of the $N$ CPFs containing $c^{(i)}$ in the $\subseteq^*$ sense.
That is, given CPFs $c^{(1)}, c^{(2)}, \ldots, c^{(N)}$, the band depth of CPF $c^{(i)}$ is $BD_N(c^{(i)}) = \binom{N}{2}^{-1} \sum_{j=1}^{N-1} \sum_{k=j+1}^N \1\big(c^{(i)} \subseteq^* B(c^{(j)}, c^{(k)})\big)$.

\subsubsection{Modified band depth}
\label{sec:mbd}

\cite{WMK13} note that if $N$ is small and the CPFs strongly vary in shape, this band depth definition can produce many zero-depth CPFs.
To circumvent this issue for one-dimensional functions, \cite{LopezPintado09} define a modified band depth that measures the proportion of a function's graph that is in the band. 
\cite{SGN12} introduce an efficient algorithm to compute this modified band depth. 
We extend this efficient algorithm to compute modified band depth for $d-$dimensional CPFs, where $d \geq 2$. 

\begin{example} 
\label{ex:depth}
\upshape
We motivate our algorithm through an illustrative example in which we compute the depth of an arbitrary CPF, denoted as $c^{(*)}$, among the $N=101$ CPFs in Figure \ref{fig:exampledepth}, where we assume $\oset = [0, 1]^2$.
First, we define and compute $c^{(*)}$'s depth at the vertical dotted line $y_1=0.8$, which we denote as $d^{(*)}|_{y_1=0.8}$.
Let $A^{(*)}|_{y_1=0.8} = \min \{y_2 \in [0, 1] : (0.8, y_2) \in A^{(*)}\}$ be the minimum $y_2$ value of the lower-left boundary of DPSC $A^{(*)}$ that intersects the vertical line $y_1=0.8$. 
Analogous to the $\subseteq^*$ relation, a pair of CPFs $c^{(j)}$ and $c^{(k)}$ is said to contain CPF $c^{(*)}$ at the line $y_1=0.8$ if and only if $A^{(*)}|_{y_1=0.8}$ is in the closed interval $[A^{(j)}|_{y_1=0.8},\, A^{(k)}|_{y_1=0.8}]$.
Then $d^{(*)}|_{y_1=0.8}$ is defined to be the fraction of pairs of the $N$ CPFs that contain $c^{(*)}$,
where the number of CPF pairs that contain $c^{(*)}$ is $|\{ i : A^{(i)}|_{y_1=0.8} \leq A^{(*)}|_{y_1=0.8} \}| \times |\{ i : A^{(i)}|_{y_1=0.8} \geq A^{(*)}|_{y_1=0.8} \}| - 1$ (we subtract one to avoid counting the band delimited by $c^{(*)}$ with itself). 
Note that $|\{ i : A^{(i)}|_{y_1=0.8} \leq A^{(*)}|_{y_1=0.8} \}|$ is simply the rank of $A^{(*)}|_{y_1=0.8}$ among $\{ A^{(i)}|_{y_1=0.8} : i = 1, \ldots, N\}$ and is also equal to  $N - |\{ i : A^{(i)}|_{y_1=0.8} \geq A^{(*)}|_{y_1=0.8} \}| + 1$.

We may repeat the process above for any vertical line, e.g. $y_1 = s$, to obtain the depth $d^{(*)}|_{y_1=s}$ of CPF $c^{(*)}$ at $y_1 = s$.
Similarly, we may easily alter the process above to obtain depth $d^{(*)}|_{y_2=t}$ of $c^{(*)}$ at any \textit{horizontal} line, e.g. $y_2 = t$.
If we create a dense uniform grid of $q$ lines for each output dimension and find the depth of $c^{(*)}$ at each of the $2q$ lines, 
we can approximate the ``overall'' depth of $c^{(*)}$ by the sample mean of these $2q$ depths: $0.5 q^{-1} \sum_{j=1}^q d^{(*)}|_{y_1=t_j} + 0.5 q^{-1} \sum_{j=1}^q d^{(*)}|_{y_2=t_j}$, where $t_j \coloneqq (j-1)/(q-1)$.
\hfill $\blacksquare$
\end{example}



Following a process similar to Example \ref{ex:depth}, Section \textbf{B.2} of the Supplement provides an explicit algorithm to find the depth of all $N$ CPFs when $d=2$.
Step 1 creates the $2q$ lines while Step 2 finds the $y_2$ and $y_1$ intersection values for all $N$ CPFs.
At each line, Step 3 ranks the CPFs while Step 4 computes for each CPF the number of pairs of CPFs that contain it. 
Step 5 then calculates the depth of each CPF. 
Section \textbf{B.3} of the Supplement extends this $d=2$ process to any $d\geq2$, which has a runtime of $\mathcal{O}(d q^{d-1} N \log N)$. 



\subsubsection{PF and PS estimation}

Our $\alpha_{MBD}\%$ PF UQ point cloud for the depth approach is the union of the $\alpha_{MBD}$ deepest CPFs, where $0 < \alpha_{MBD} < 1$.
Our PS UQ region is then the union of the hyperrectangles corresponding to the points in the PF UQ point cloud.

\subsection{Comparing time complexity}
\label{sec:runtime}

To compare the runtime of the two approaches for any $d \geq 2$,
we first express the random-sets runtime as $\mathcal{O}(d N^2 \bar{n}^2)$, where $\bar{n} \coloneqq N^{-1} \sum_{i=1}^N |c^{(i)}|$ is the mean number of points in each CPF.
Inference can be controlled by varying either the number of posterior draws ($N$) or the training size ($n$).
If we fix $N$ and increase $n$ (and hence also $\bar{n}$), the random-sets runtime grows quadratically in $\bar{n}$,
but we would also want to grow $q$ proportionally to $\bar{n}$ in order to faithfully capture the ranks of the increasingly refined CPFs, which would affect the depth runtime via $q^{d-1}$.
Hence, the comparison between the two runtimes depends on $d$. 
In practice, however, the training size is fixed (which also roughly fixes $\bar{n}$).
As $N$ increases, the depth runtime grows more slowly than does the random-sets runtime ($N \log N$ vs $N^2$), which makes the depth approach the more computationally tractable option in this case.
\section{Simulation study}
\label{sec:simstudy}

\paragraph{Simulation settings.} 
We generate data from one of four test functions: MOP2 \citep{Fonseca1995}, ZDT3 \citep{Zitzler00}, DTLZ2 \citep{Deb2005}, and ZLT1 \citep{Laumanns2005}, which for brevity are defined in the Supplement. 
Figure \ref{fig:MOP2ZDT3} shows the MOP2 function ($p=d=2$) to be the simplest of the four. 
ZDT3 ($p=d=2$) has a disconnected PF and PS. 
DTLZ2's ($p=4$ and $d=2$) PF and PS (not shown) are similar to MOP2's.
ZLT1's ($p=d=3$) PF is a convex 2-dimensional surface while its PS is the 2-dimensional probability simplex $\{(x_1,x_2,x_3) \in [0,1]^3: x_1+x_2+x_3=1\}$.
Though the methodology in Sections \ref{sec:bartpareto} and \ref{sec:uq} is invariant to shifts and scales of inputs or outputs, our performance metrics in Section \ref{sec:perfmet} are not. 
Thus, we shift and scale the input space to be $\iset = [0, 1]^p$ and each objective to have range $[0, 1]$.

\begin{figure}[t!]
    \centering
    \includegraphics[width=0.24\textwidth]{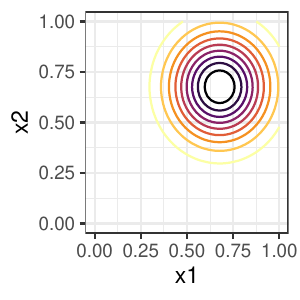}
    \includegraphics[width=0.24\textwidth]{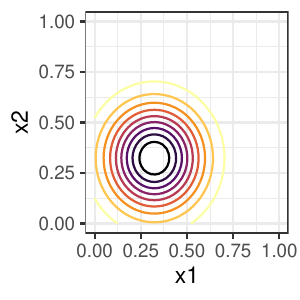}
    \includegraphics[width=0.24\textwidth]{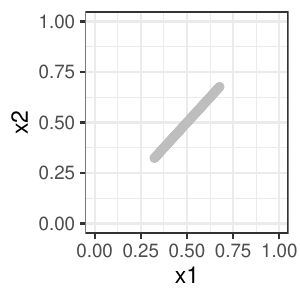}
    \includegraphics[width=0.24\textwidth]{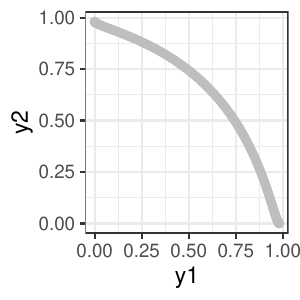}
    \includegraphics[width=0.24\textwidth]{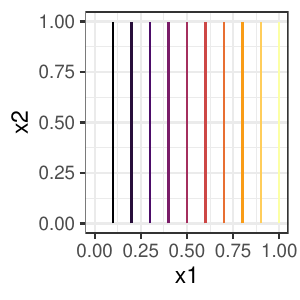}
    \includegraphics[width=0.24\textwidth]{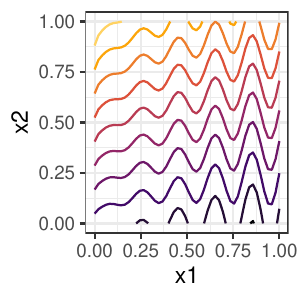}
    \includegraphics[width=0.24\textwidth]{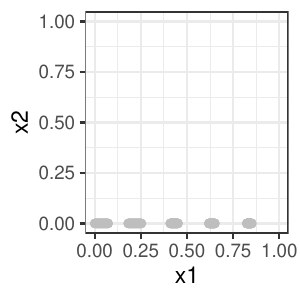}
    \includegraphics[width=0.24\textwidth]{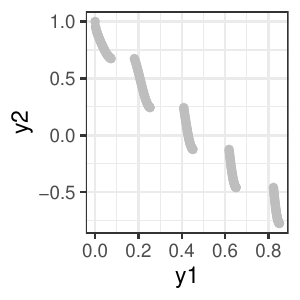}
    \includegraphics[width=0.24\textwidth]{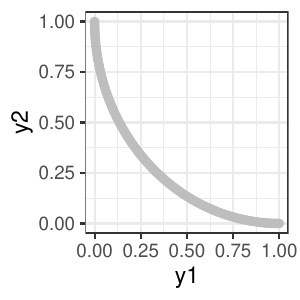}
    \includegraphics[width=0.31\textwidth]{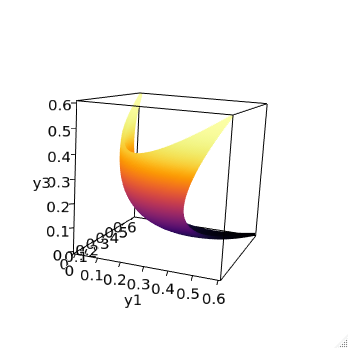}
    \caption{The first row (MOP2) and second row (ZDT3) show, from left to right, the respective function's $f_1$ contour, $f_2$ contour, PS, and PF. 
    For these plots, darker contour lines indicate lower objective values.
    The third row shows the PF of DTLZ2 (left) and ZLT1 (right), where darker values indicate lower $y_2$ values.
    }
    \label{fig:MOP2ZDT3}
\end{figure}

Given data from \eqref{eq:datagen} and $\f(\bm{\cdot})$'s PF and PS, this section explores how sample size and measurement error magnitude affect the accuracy of:
\begin{enumerate}[label=\textbf{Q.\arabic*}]
    \item \label{qu:pf} BART's PF point clouds and GP's PF approximate point clouds.
    \item \label{qu:ps} BART's PS point clouds.  
    \item \label{qu:uq} Depth approach and random sets approach to uncertainty quantification.
\end{enumerate}
For each test function $\f(\bm{\cdot})$, we explore the six possible combinations of the following parameter settings: noise-variance multiplier $\zeta \in \{0.0, \, 0.1, \, 0.25\}$ and training size $n \in \{32p, 64p\}$.
Given any $\left(\f(\bm{\cdot}), n, \zeta \right)$ combination,  
we can create a data set $\dat \coloneqq \{(\y(\x_i), \x_i)\}_{i=1}^n$ by simulating  $\y(\x_i)$ from \eqref{eq:datagen} at each design point $\x_i$ of an $n-$point maximin LHS on $[0, 1]^p$ \citep[from][]{spacefillingdesigns}, where $\sigma^2_j = \zeta \Var(f_j(\X))$ for all $j=1, \ldots, d$ and $\X = (X_1, \ldots, X_p)$ with each $X_i \iid U(0, 1)$.
For each $\left(\f(\bm{\cdot}), n, \zeta \right)$ combination,  
we generate either $100$ such data sets $\dat \coloneqq \{(\y(\x_i), \x_i)\}_{i=1}^n$ if $\f(\bm{\cdot})$ is one of the three $d=2$ functions, or $50$ such data sets if $\f(\bm{\cdot})$ is ZLT1. 
To each data set, we fit two models: a multiple-output BART model and a multiple-output GP model produced by fitting an independent single-output GP to each marginal data set $\dat_j$ for $j=1, \ldots, d$. 
For each model our BTE (burn-in $B$ steps, terminate after $T$ steps, sample every $E$ steps) is $(300, 800, 1)$, resulting in $500$ posterior draws per generated data set.
To summarize, we fit a multiple-output BART model and a multiple-output GP model to each of the $2,100$ data sets, obtain $500$ posterior draws from each of the $4,200$ models, find or approximate the image at each posterior draw, find the CPF of each image, compute the depth and empirical attainment function value for each CPF to produce PF (and PS if using a BART model) UQ point clouds, and compute performance metrics (defined in Section \ref{sec:perfmet}) for each point cloud.  
Running the simulation study pipeline for ZLT1 (where $d=3$) for 50 data sets took roughly 6 days with 32 cores (approximately $8.3$ hours per BART model run). 
However, we note the much shorter runtime for the engineering application in Section \ref{sec:engapp}, where $d=2$. 

We use the OpenBT implementation of BART \citep{openbt} with default parameter settings unless otherwise stated in Section \ref{sec:bart}. 
For the GP method, we use the \texttt{km} and \texttt{simulate} functions of \cite{Roustant12}'s DiceKriging package with error variance set to the scenario's noise variance.
Because noise variance is not known in most applications, the GP fits can be seen as idealized. 
For UQ, we use $\alpha_{RS} = 0.25$ and $\alpha_{MBD} = 0.5$.

\subsection{Performance metrics}
\label{sec:perfmet}

\begin{table}[h!]
  \begin{center}
    \begin{tabular}{| c | c | c | c |}
    \hline
    undercoverage & overcoverage & biased coverage & good coverage \\ 
      \includegraphics[width=1.4in]{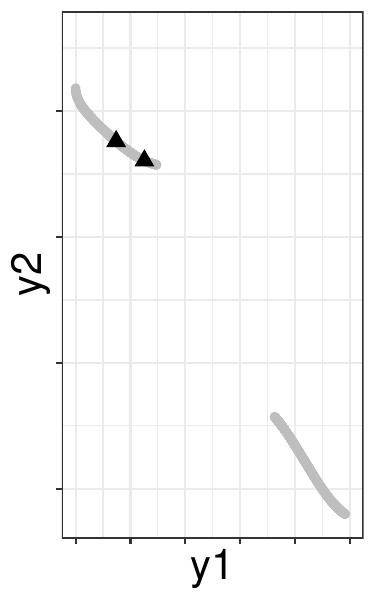} & 
      \includegraphics[width=1.4in]{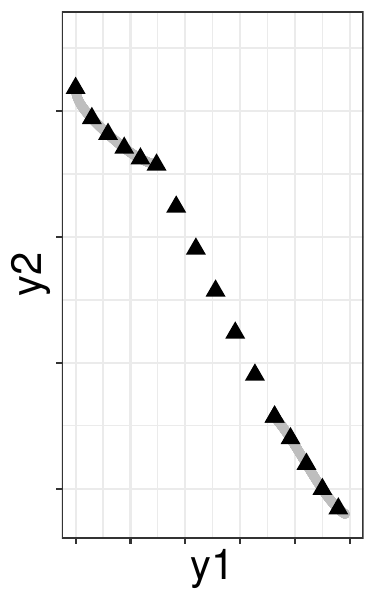} & 
      \includegraphics[width=1.4in]{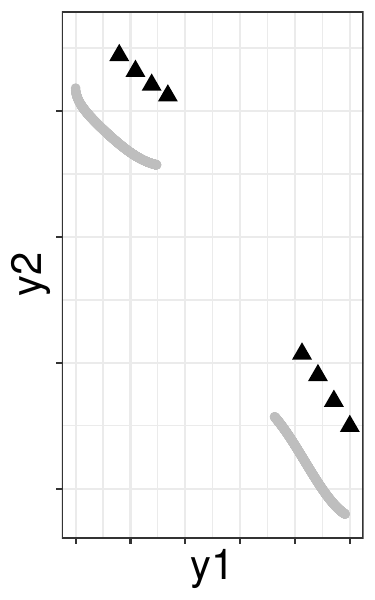} & 
      \includegraphics[width=1.4in]{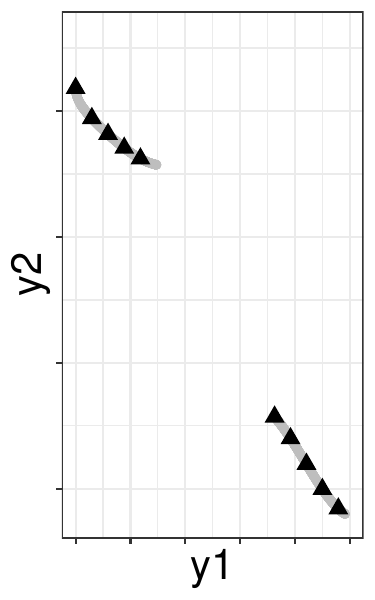} \\ \hline 
    small $d_1$, large $d_2$ & large $d_1$, small $d_2$ & large $d_1$, large $d_2$ & small $d_1$, small $d_2$ \\ \hline
    \end{tabular}
  \caption{Each plot shows a different PF point cloud (black triangles) that attempts to capture the same, disconnected target set (gray lines). Of the four point clouds, the one in the fourth panel is the only one that adequately captures target set. Qualitative values of $d_1$ and $d_2$ for each point cloud are provided.}
  \label{tbl:coverage}
  \end{center}
\end{table}

This section defines two performance metrics that jointly quantify how well a point cloud $\phat$ (either $\pfhat$ or $\pshat$) estimates its target set $\ptrue$ (either $\pftrue$ or $\pstrue$):
\begin{equation*}
d_1 \coloneqq d(\phat, \ptrue) \quad \text{and} \quad d_2 \coloneqq d(\ptrue, \phat) 
\end{equation*}
where $d(\mathcal{A}, \mathcal{B}) \coloneqq |\mathcal{A}|^{-1} \sum_{\mathbf{a} \in \mathcal{A}} (\min_{\mathbf{b} \in \mathcal{B}} \|\mathbf{a} - \mathbf{b}\|_2)$ is the average distance from points in a finite point set $\mathcal{A}$ to a set $\mathcal{B}$ \citep{Dubuisson94}.
These two metrics are analogous to the Type I / Type II error of a hypothesis test: a point cloud with many points far from the target set (similar to false negatives) will have a large $d_1$ value while a point cloud far away from many points in the target set (similar to false positives) will have a large $d_2$ value. 
That is, the metrics $d_1$ and $d_2$ measure the degree to which a point cloud exhibits these two undesirable behaviors.
As examples, the point clouds in the first and third panels of Table \ref{tbl:coverage} have large $d_1$ values while the point clouds in the second and third panels have large $d_2$ values.
Conversely, a point cloud with small $d_1$ and $d_2$ values, e.g. in the fourth panel, indicates it is a high-performing estimate of the target set.

Our metric $d_1$ is equivalent to the function $\mathcal{M}_1$ in \cite{Zitzler00}, which is one of 63 performance indicators in the MO literature reviewed by \cite{Audet2020}. 
Though many of these indicators penalize only one of the two mentioned undesirable behaviors, some more recent indicators penalize both behaviors with a single metric to be used in a sequential MO design. 
This paper, however, focuses on characterizing the performance of estimates of $\pftrue$ and $\pstrue$.
Hence, we penalize these two behaviors separately to see \textit{how} a point cloud might underperform.

\subsection{Simulation results}
\label{sec:simres}

\begin{figure}
    \centering
    \includegraphics[width=\textwidth]{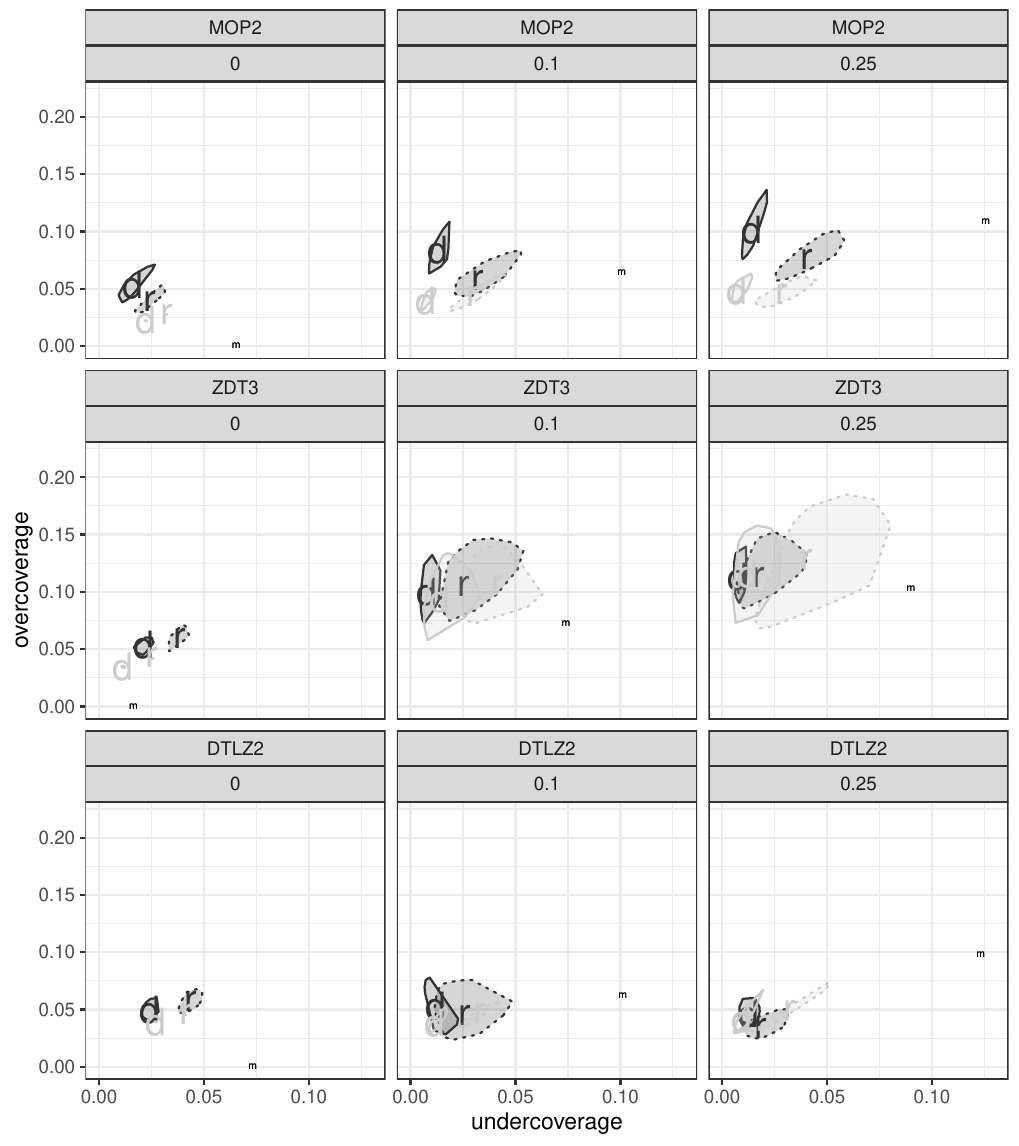} 
    \caption{Bagplots of the 100 values of $d(\pfhat, \pftrue)$ and $d(\pftrue, \pfhat)$. Each plot represents a $n = 64p$ simulation scenario where the variance multiplier is a value in $\{0, 0.1, 0.25\}$. Bags with solid (dashed) outline and median labeled `d' (`r') display depth (random sets) approach. Black (gray) bags display BART (GP) model.}
    \label{fig:resultsPF64p}
\end{figure}

Figure \ref{fig:resultsPF64p} displays bagplots of the 100 values of $d(\pfhat, \pftrue)$ and $d(\pftrue, \pfhat)$ for each $n=64p$ simulation scenario (PS plots and similar $n=32p$ scenario results can be found in the Supplement).
A bagplot extends the common boxplot for two-dimensional outputs and contains three main features analogous to the common univariate median, the box, and the whiskers on a conventional boxplot \citep{Rousseeuw1999}. 
For visual clarity, we include only two of these features: the depth median, which is the point with the highest possible halfspace depth, and the ``bag'', which is a polygon that encloses 50\% of the points around the depth median. 

As another point of comparison, we also plot in Figure \ref{fig:resultsPF64p} the median undercoverage and overcoverage of 100 sets of 10 points randomly selected from the underlying function's PF and perturbed according to the scenario's noise level. 
These medians are the points labeled `m'.
Each set of 10 randomly selected points from $\pftrue$ provides a sense of the undercoverage and overcoverage one can expect from a ``well-fitting'' statistical model that produces a 10-point CPF. 
The subsequent perturbation accounts for the quality of the data that a statistical model is trained on.
We do not perform this comparison for the PS figures for lack of a natural mapping of observation noise level to input noise level.

To address \ref{qu:pf}, we make several observations in Figure \ref{fig:resultsPF64p}. 
We first compare the difficulty of MO between the three $d=2$ functions. 
The ZDT3 PF point clouds tend to produce higher overcoverage than either the MOP2 or DTLZ2 PF point clouds, which can be explained by ZDT3's disconnected PF. 
The PF point cloud undercoverage is roughly the same between the three $d=2$ functions. 
The ZLT1 PF point cloud (Supplement, Figures 5a), 
however, produces much more undercoverage than those of the other three functions, presumably because 2-dimensional surfaces (e.g. ZLT1's PF) are usually more difficult to cover than 1-dimensional sets (e.g. the other three PFs). 
However, the PS point clouds (Supplement, Figures 2b, 3b, 4b, and 5b) perform differently between the three $d=2$ functions. 
When $d=2$, the DTLZ2 PS point clouds have the most undercoverage, which may be due to a larger input dimension ($p=4$).
These point clouds also have the most overcoverage (when $d=2$), but the overcoverage difference between DTLZ2 and ZDT3 is roughly the same as the overcoverage difference between ZDT3 and MOP2, which again can be explained by ZDT3's PS being disconnected and on the boundary of the input space. 
Similar to the PF regime, the ZLT1 PS point cloud produces much more undercoverage than those of the other three functions, which again may be explained by ZLT1's $2-$dimensional PS.
Surprisingly, the ZLT1 PS point clouds have fairly low overcoverage, which suggests we could increase $\alpha_{MBD}$ and $\alpha_{RS}$ to improve undercoverage. 
From these observations, we conclude that among the $d=2$ functions, ZDT3 has the most difficult PF to capture while DTLZ2 has the most difficult PS to capture, but the increased dimensionality of ZLT1's PF and PS makes point cloud prone to have large undercoverage. 

We also compare performance between BART and (idealized) GP. 
In the MOP2 PF results, the no-noise scenario shows an overcoverage/undercoverage tradeoff between BART and GP while the two noisy $n=128$ scenarios show GP outperforming BART in both metrics.
The DTLZ2 PF results show roughly equal performance between BART and GP. 
The ZDT3 PF results show BART's performance improving relative to GP's performance as noise increases, which suggests that the fitted stationary GP models struggle with ZDT3's irregular oscillations in its image.
These observations imply that BART performs possibly worse than idealized GP (which presumably performs better than a fitted GP when noise is not known) in ``simpler'' scenarios but better adapts to ``complex'' behaviors in the underlying data-generating function.
We conclude that when the underlying function and noise level are not known, BART may be a safer bet than the GP. 

To address \ref{qu:ps}, 
we refer to Figures 2b, 3b, 4b, and 5b in the Supplement. 
For each $d=2$ test function, the depth approach tends to produce similar overcoverage and undercoverage as the random sets approach. 
Overcoverage tends to be larger than undercoverage for each function and each approach, which suggests $\alpha_{MBD}$ and $\alpha_{RS}$ could be lowered to produce point clouds with less overcoverage and minimally more undercoverage. 
For ZLT1, the difference between the two approaches is slightly more pronounced. 
When $n=96$, there seems to be a tradeoff between overcoverage and undercoverage, but when $n=192$, the depth approach seems to have slightly less undercoverage and roughly equal overcoverage as the random sets approach. 
Interestingly, undercoverage of ZLT1's PS point clouds remains large and does not decrease with larger sample size, which suggests there is a region of the PS that the point clouds consistently fail to cover. 

To address \ref{qu:uq}, we now compare the depth approach to the random sets approach. 
We first look at BART's PF point clouds.
For the MOP2 and ZLT1 functions, the depth approach tends to produce more overcoverage and less undercoverage than the random sets approach. 
For ZDT3, the depth approach tends to produce less overcoverage and undercoverage than the random sets approach.
For DTLZ2, the depth approach tends to produce less overcoverage and undercoverage than the random sets approach when the observations are not noisy, but more overcoverage when noise is present. 
In all of these BART PF observations, the depth approach either outperforms or produces an overcoverage/undercoverage tradeoff with the random sets approach. 
That is, in no BART PF scenario does the random sets approach outperform the depth approach, which suggests the depth approach produces either as good or better PF point clouds than does the random sets approach.
For GP, the depth approach tends to produce less overcoverage and less undercoverage than the random sets approach for all three $d=2$ test functions, which suggests the depth approach produces overall better GP-based PF point clouds than does the random sets approach. 
From these observations and from the depth approach's runtime advantage as discussed in Section \ref{sec:runtime}, 
we conclude that the depth approach should be used over the random sets approach if using either BART or GP.

\section{Engineering application}
\label{sec:engapp}

Consider the single cut turning cost operation from \cite{Trautmann2009} (TM), who consider the MO problem of simultaneously minimizing the machining and tool costs,  $C_m \colon \iset \rightarrow \mathbb{R}$ and $C_t \colon \iset \rightarrow \mathbb{R}$, respectively, for an industrial engineering application where
\begin{align}
  C_m(\nu_c, f_r) &= b_1 \nu_c^{-1} f_r^{-1} \quad \text{and} \quad C_t(\nu_c, f_r) = b_2 \nu_c^2 f_r^3
  \label{eq:MC}
\end{align}
with constants $b_1 \approx 12,354 \, \pounds \text{ mm}^2 \text{ min}^{-1}$ and $b_2 \approx 0.0284 \, \pounds \text{ mm}^{-5} \text{ min}^{2}$.
Each cost has two input variables: cutting speed $\nu_c$ with typical values between 10 and 400 mm/min, and the feed $f_r$ with typical values between 0.04 and 1 mm.
Thus, we use the input space $\iset = [10, 400] \times [0.04, 1]$.

\begin{figure}[h!]
    \centering
    \includegraphics[width=0.32\textwidth]{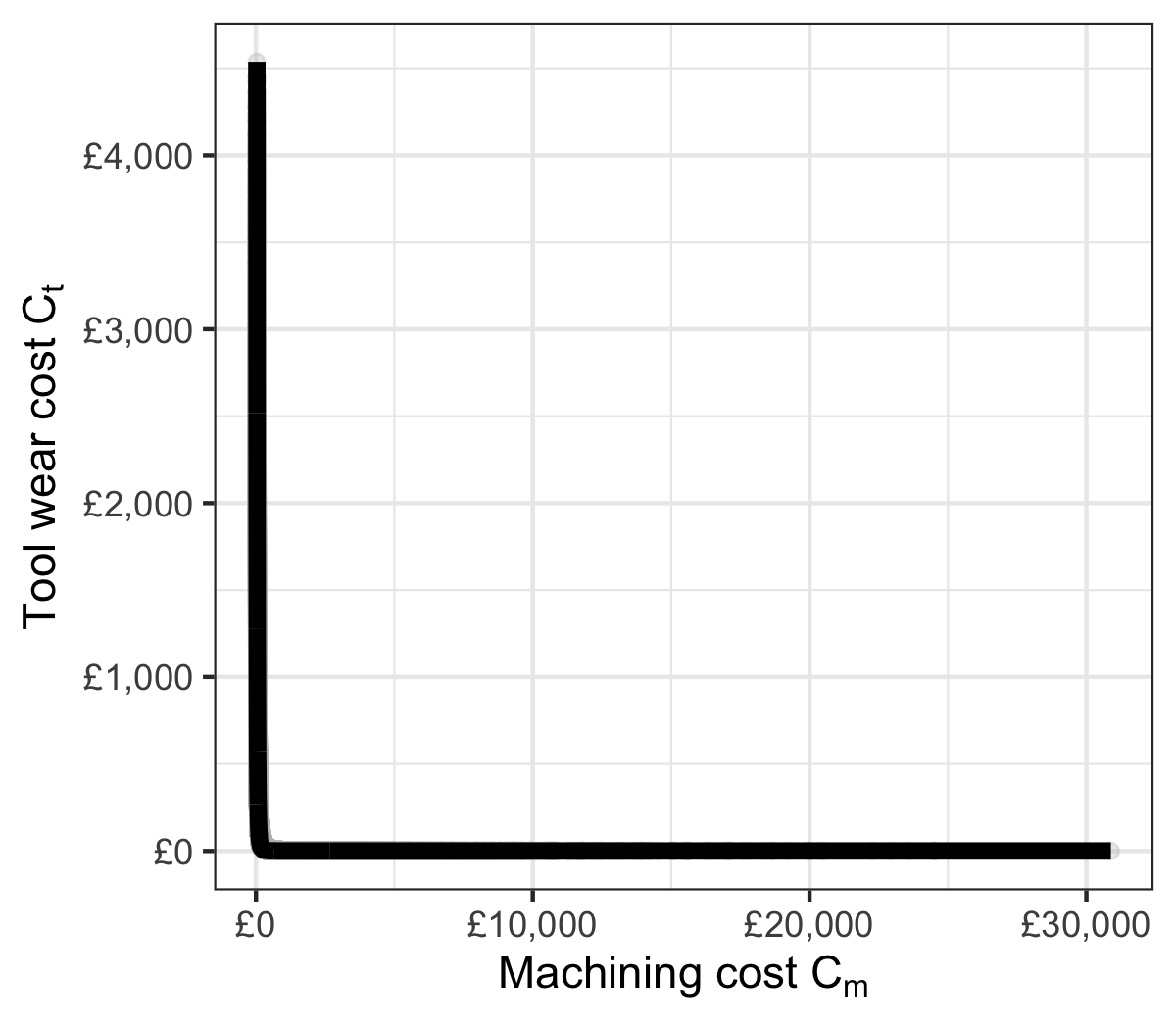} 
    \includegraphics[width=0.32\textwidth]{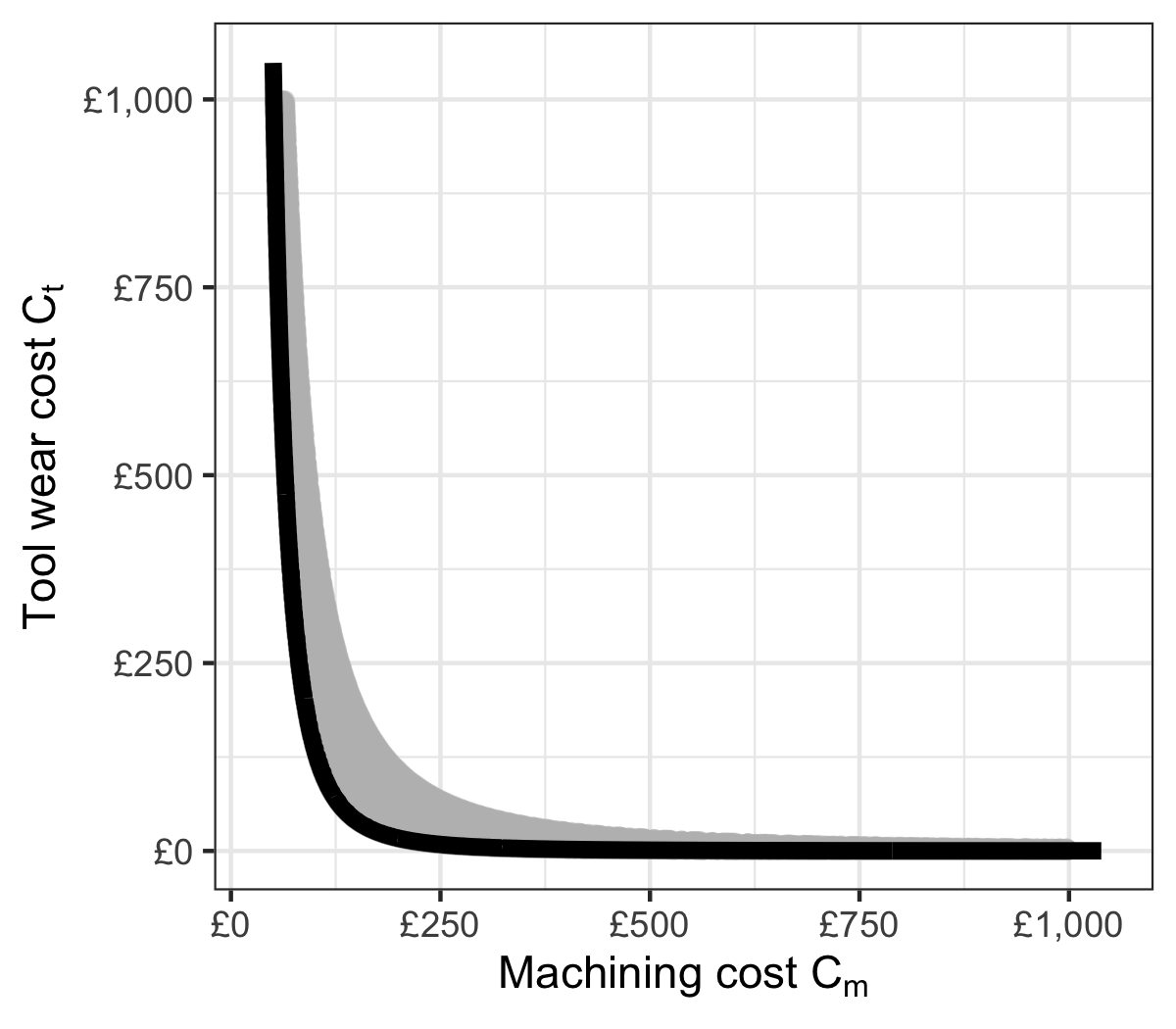} 
    \includegraphics[width=0.32\textwidth]{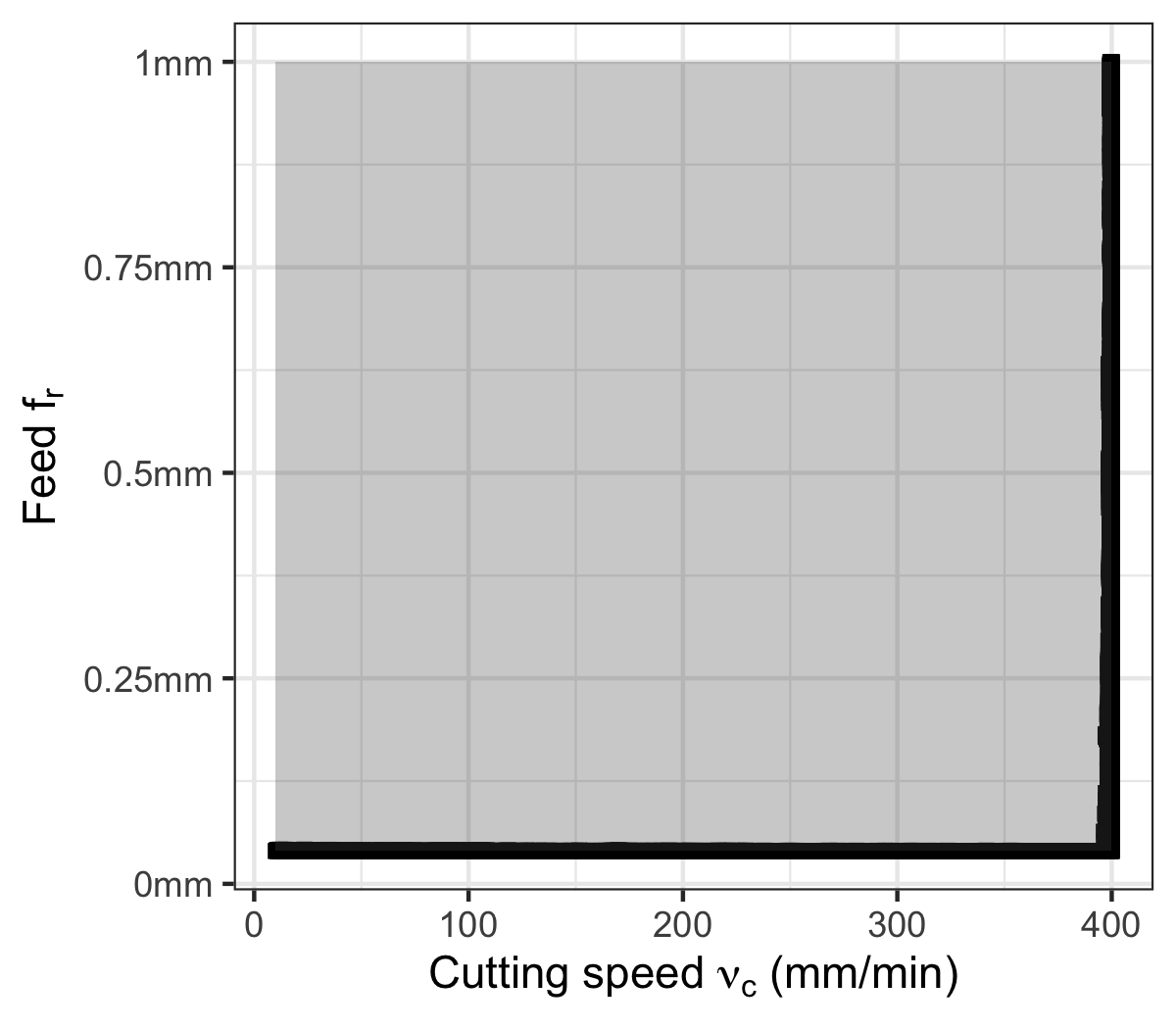} 
    \caption{Left: image (gray) and PF (black) of the single cut turning cost operation biobjective function. Middle: the left plot zoomed into the lower-left region. Right: input space (gray) and PS (black).}
    \label{fig:MCPFPS}
\end{figure}

As shown in the left and middle plots of Figure \ref{fig:MCPFPS}, no image point is far from the PF. 
Thus, it is ``easy'' to estimate the PF.
However, this same property in the output space makes PS estimation very difficult, as any input will map to an image point close to or on the PF.
Because many input points are far from the PS (shown in the right plot of Figure \ref{fig:MCPFPS}), PS estimates will tend to have large uncertainty.


\begin{figure}[h!]
    \centering
    \begin{subfigure}[t]{.23\textwidth}
        \centering
        \includegraphics[width=\textwidth]{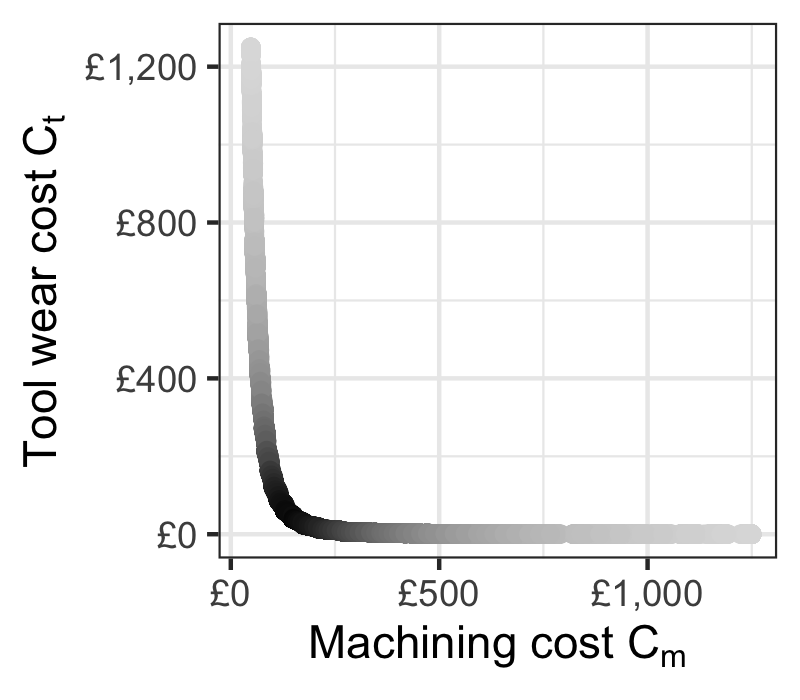} 
        \caption{$n = 15300$ PF UQ.}
        \label{fig:MCpf15300}
    \end{subfigure}
    \begin{subfigure}[t]{.23\textwidth}
        \centering
        \includegraphics[width=\textwidth]{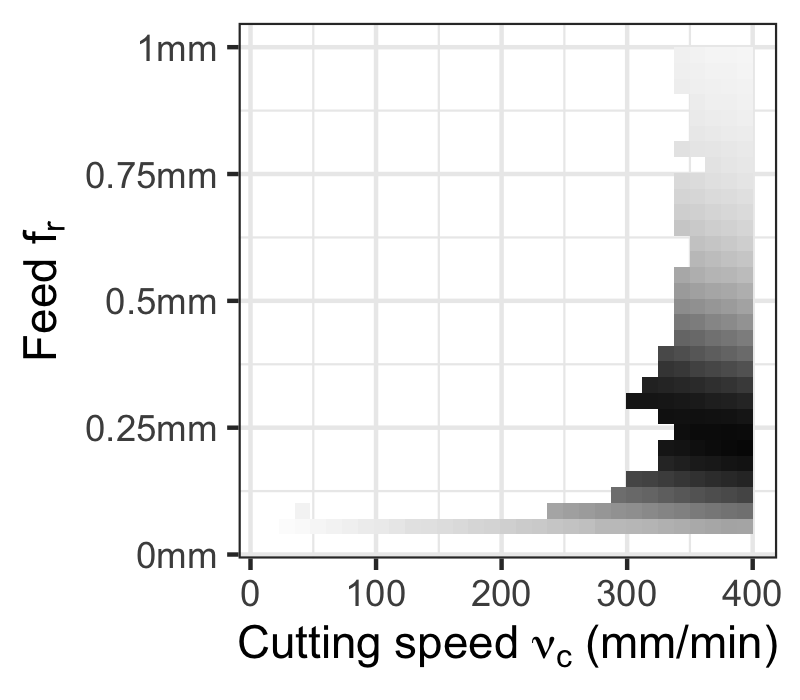} 
        \caption{$n = 15300$ PS UQ.}
        \label{fig:MCps15300}
    \end{subfigure}
    ~
    \begin{subfigure}[t]{.23\textwidth}
        \centering
        \includegraphics[width=\textwidth]{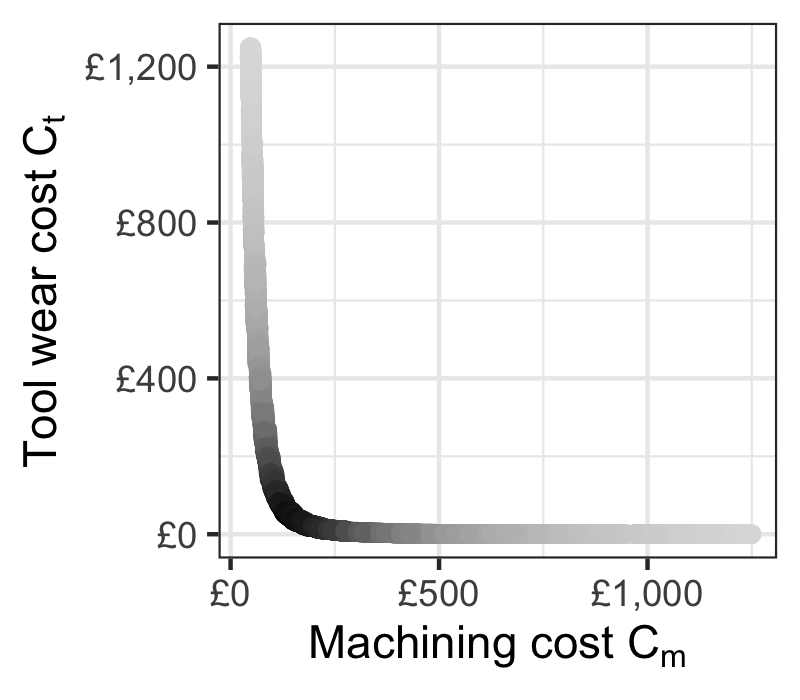} 
        \caption{$n = 1500$ PF UQ.}
        \label{fig:MCpf1500}
    \end{subfigure}
    \begin{subfigure}[t]{.23\textwidth}
        \centering
        \includegraphics[width=\textwidth]{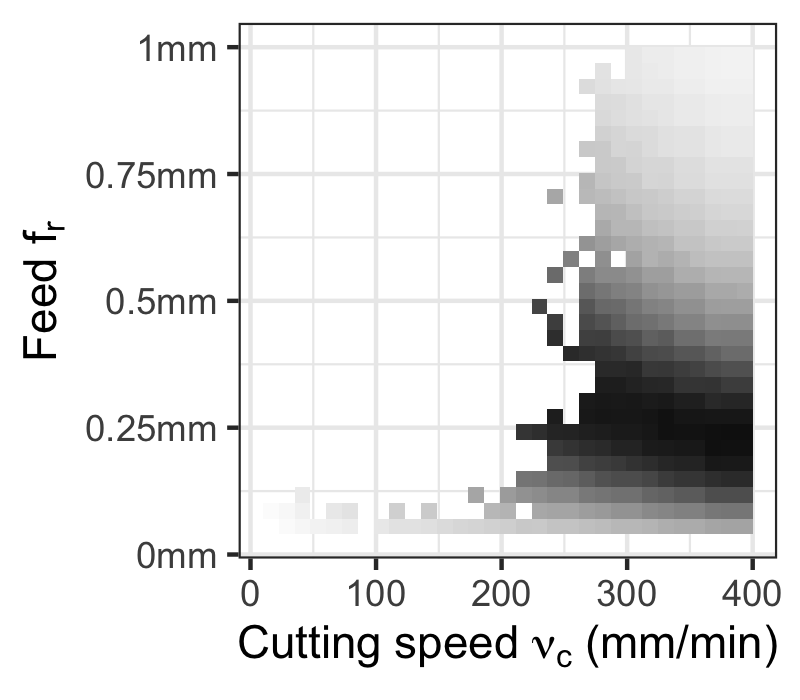} 
        \caption{$n = 1500$ PS UQ.}
        \label{fig:MCps1500}
    \end{subfigure}
    \caption{50\% UQ point clouds for the PF and PS.}
    \label{fig:resultsMachiningCost}
\end{figure}

TM perform PF estimation but not UQ or PS estimation, which hides the aforementioned large uncertainty in PS estimates.
In contrast, we apply our BART-based methodology (with BTE $(300,2300,1)$) to perform PF/PS estimation and UQ on two training data sets generated from \eqref{eq:datagen}, where $\sigma^2=0$ and \eqref{eq:MC} takes the role of $\f(\bm{\cdot})$.
We generate noiseless data, but approach the problem as if we do not \textit{a priori} know the level of noise in the observations.
Due to sharp peaks in $C_m$ and $C_t$ in the generated data, we fit our two BART models to $\log C_m$ and $\log C_t$. 
We then transform the BART predictions back to $C_m$ and $C_t$ before performing PF/PS estimation and UQ.
For UQ, we use only the depth approach per our simulation study conclusions in Section \ref{sec:simres}. 
The runtime of this analysis for $n = 15300$ was 2 minutes and 45 seconds with 32 cores.

Figure \ref{fig:resultsMachiningCost} shows the $50\%$ PF and PS point clouds for two training sizes: $n=15300$ to match the number of function evaluations made by TM, and $n=1500$ to consider the case of an expensive simulator.
The dark PF points correspond to the dark PS regions and indicate relatively low $C_m$ and $C_t$ values. 
As explained in the previous paragraph, both PS point clouds (Figs. \ref{fig:MCps15300} and \ref{fig:MCps1500}) show large uncertainty. 
On the other hand, the two PF point clouds (Figs. \ref{fig:MCpf15300} and \ref{fig:MCpf1500}) indicate small uncertainty and differ only slightly from each other, which implies only a minor loss in PF inference even with a training size reduction of $90\%$.
Hence, a practitioner can pick an input setting to achieve relatively low $C_m$ and $C_t$ values even with $n=1500$.

If such a single cut turning cost operation is embedded in a continuously monitored and adapting manufacturing system, it makes sense to allow cutting speed to be continuously manipulated by automated systems using relevant manufacturing information obtained in real-time.
For example, a sudden jump in the price of raw materials might increase the impact of cutting speed on tool wear cost and hence change the nature of the PF. 
Our methodology provides an automated approach to understanding which input settings will result in low-cost outcomes.

\section{Summary and Discussion}
\label{sec:conc}

Using the fact that BART produces a set of hyperrectangles which partition the input domain (Theorem \ref{thm:multibartimg}), this paper describes the details of using BART for performing multiobjective optimization, provides an algorithm to find the PF and PS of the multiple outputs and inputs, and compares two different approaches of UQ for the PF and PS. 
A ``random-sets'' approach and a newly proposed ``depth'' approach are used to quantify the uncertainty of BART-generated and GP-generated PF and PS estimates.
The depth approach performs similarly or better than the random sets approach (while being  computationally advantageous).
When the underlying function and noise level is unknown, UQ based on BART-based MO optimization may be superior to GP-based MO optimization. 
We also note that BART can readily handle categorical inputs, which are often a challenge in GP models.
Finally, we demonstrated our BART-based PF and PS estimation to data generated from an engineering application. 


This paper suggests several topics for additional research.  
First, our UQ comparisons used $\alpha_{MBD} = 0.50$ and $\alpha_{RS} = 0.25$, but we could lower these values to decrease the expected overcoverage and increase the expected undercoverage (or raise these $\alpha$ values to increase expected overcoverage and decrease expected undercoverage). 
Indeed, the empirical results in this paper suggest using lower values than our $\alpha$ choices  but it is unknown what $\alpha_{MBD}$ and $\alpha_{RS}$ values will produce desirable overcoverage/undercoverage values in general.


Second, this paper used one-stage maximin LHDs. 
Presumably, BART's PF and PS estimates could be improved using alternative input designs. 
For example, \cite{Chipman12} performed sequential design using single-output BART prediction, but additional design research is required for multiple-output BART. 

Third, this paper assumes an independent correlation structure in the distribution of $\bm{\epsilon}_i$ in \eqref{eq:datagen}. 
Introducing dependence would require a modeling approach that cannot be accomplished via independent BART model fits. 
To date, no such approach exists.

The implementation for finding the PF and PS of a two-output or three-output BART model can be found in the Open Bayesian Trees (OpenBT) project at \url{https://bitbucket.org/mpratola/openbt/}.


\bigskip
\begin{center}
{\large\bf ACKNOWLEDGEMENTS}
\end{center}

\if0\blind
{
    AH~would like to acknowledge the Graduate School at The Ohio State University for support during the dissertation year. 
    TJS~was supported in part by the National Science Foundation under Agreements  
    DMS-0806134 and DMS-1310294 (The Ohio State University).
    YS~was supported by King Abdullah University of Science and Technology (KAUST), Oﬃce of Sponsored Research (OSR) under Award No: OSR-2019-CRG7-3800.
    The work of MTP~was supported in part by the National Science Foundation under Agreement DMS-1916231 and in part by the King Abdullah University of Science and Technology (KAUST) Office of Sponsored Research (OSR) under Award No. OSR-2018-CRG7-3800.3.
} \fi

\newpage

\bigskip
\begin{center}
{\large\bf SUPPLEMENTARY MATERIAL}
\end{center}


\begin{description}

\item[Proof, Algorithms, Function definitions, Additional simulation results] 
Proof of Theorem \ref{thm:multibartimg}.
Pseudocode for \cite{Kung75}'s algorithm to find the PF of a set of $d-$dimensional vectors as described in Section \ref{sec:bartpareto}.
Description of the algorithm to compute modified band depth as described in Section \ref{sec:mbd}.
Definitions of the test functions introduced in Section \ref{sec:simstudy}.
ZLT1 simulation study plots, PS plots, and $n=32p$ scenario results as described in Section \ref{sec:simstudy}. 
(pdf file)






\item[R code for Section \ref{sec:engapp}:] R code for Section \ref{sec:engapp}. (zip file)


\end{description}


\bibliographystyle{Chicago}

\bibliography{bart}

\end{document}